%% file: deep_analog_cvpr.tex
\pdfoutput=1  
\documentclass[10pt,twocolumn,letterpaper]{article}
\usepackage[pagenumbers]{cvpr}

\usepackage{graphicx}
\usepackage{amsmath}
\usepackage{amssymb}
\usepackage{booktabs}
\usepackage{array}
\usepackage{multirow}
\usepackage{subcaption}
\usepackage{enumitem}
\usepackage{float}
\usepackage{pifont}


\definecolor{cvprblue}{rgb}{0.21,0.49,0.74}
\usepackage[pagebackref,breaklinks,colorlinks,allcolors=cvprblue]{hyperref}

\renewcommand{\ie}{i.e.}
\renewcommand{\etal}{et al.}
\newcommand{\R}{\mathbb{R}}

\begin{document}

\title{Deep Analog: Open-Set Film Emulation with Reference-Conditioned 3D LUTs}
\author{Yitong Mu\\
Rochester Institute of Technology\\
Rochester, NY 14623\\
{\tt\small ym5445@rit.edu}
}

\twocolumn[{
  \renewcommand\twocolumn[1][]{#1}
  \maketitle
  \vspace{-0.5em}
  \begin{center}
    \includegraphics[width=\textwidth]{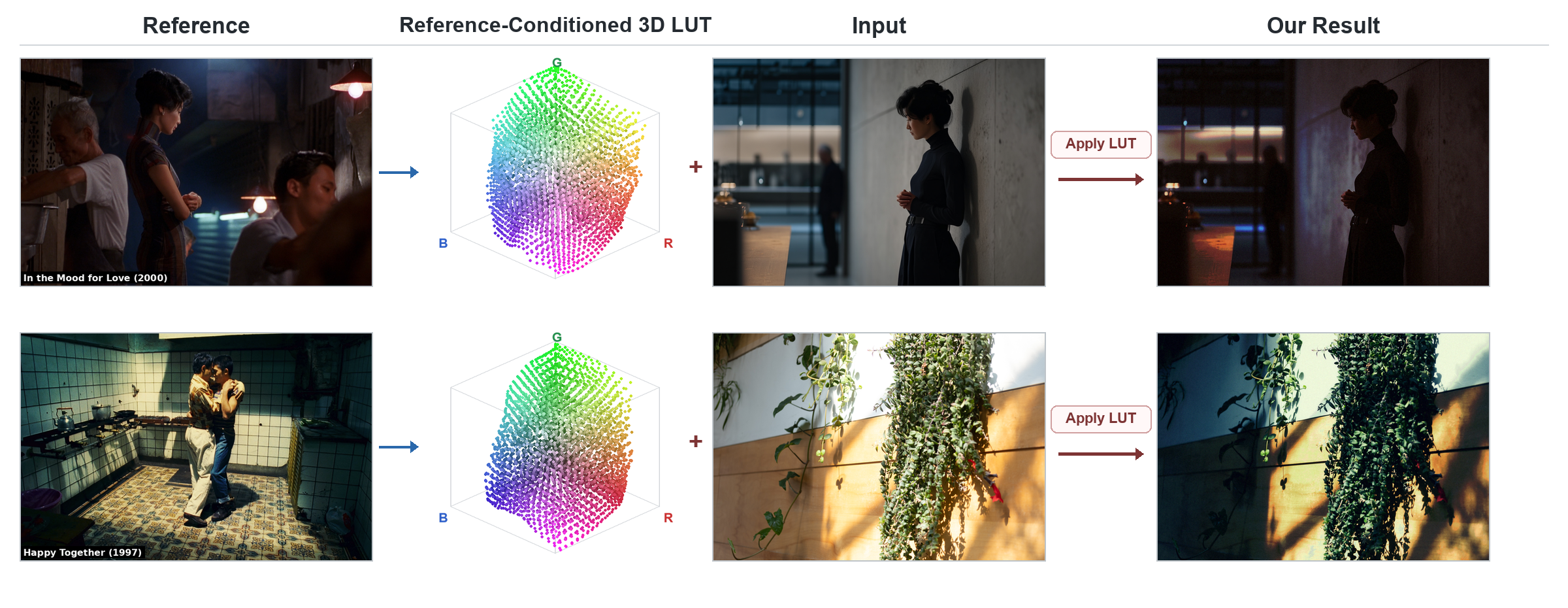}
    \captionof{figure}{Open-set film look transfer with Deep Analog. Each row shows a reference film frame, its reference-conditioned 3D LUT, an arbitrary digital input, and the final output after applying the inferred look. Point positions and colors in the LUT visualization encode output RGB coordinates. Top: the warm, low-key palette of \textit{In the Mood for Love} (2000). Bottom: the teal-and-gold cross-processed palette of \textit{Happy Together} (1997). The same model handles both references without retraining or manual parameter tuning.}
    \label{fig:teaser}
  \end{center}
  \vspace{0.5em}
}]

\begin{abstract}
Film emulation reproduces the look of an analog film stock on a new digital photograph.
We target its open-set form---matching the look of \emph{any} reference film frame from a single example---with a 3D lookup table (LUT) predicted from that reference.
Real-time image enhancement has converged on one recipe for the color step: predict per-image weights over a fixed bank of 3D LUTs and blend them.
We show this construction is a gated mixture of experts---specialist tables chosen per image by a learned gate---and it inherits the corresponding failure: trained end-to-end against reconstruction, the gate collapses onto a single expert, so a bank of $K$ LUTs delivers the capacity of one.
In a five-LUT model, one table draws 95\% of the weight and four sit idle (mean gate entropy 0.16 of a possible 1.61).
An entropy term, the enhancement-setting analogue of mixture-of-experts load balancing, restores utilization and recovers about 1~dB PSNR.
The deeper constraint survives the fix: a fixed LUT basis is closed-set, freezing the achievable looks at training time.
We therefore discard the basis and predict a single 3D LUT as a residual from a reference image (\textit{StyleLUTNet}), trained by self-supervision on procedurally generated color transforms.
The conditional design removes the gate and generalizes open-set to arbitrary, unseen film stocks without paired data or retraining.
Around this color backbone we build \textit{Deep Analog}, a film-emulation pipeline that adds histogram-based tone matching and a physics-informed optical renderer---multi-scale grain and per-channel halation driven by parameters an inverse network regresses from the reference.
On 350 self-supervised pairs the color stage reaches 22.05~dB PSNR / 0.925 SSIM and the full pipeline 21.72~dB / 0.923; the color path runs in 5.2~ms at 1080p (192~FPS) and exports a portable \texttt{.cube} LUT for Photoshop, DaVinci Resolve, and Lightroom.
A second degeneracy in conditional LUT training---residual-scale collapse---shares the root cause and yields a general operating principle: auxiliary regularization must stay subordinate to reconstruction.
\end{abstract}

\section{Introduction}
\label{sec:intro}

Digital photography dominates contemporary image capture, yet persistent demand exists for the visual character of analog film---its non-linear tonal rolloff, grain structure, and light-scattering artifacts such as halation, the soft glow that spreads around bright highlights.
Software presets that approximate these qualities have become a commodity in creative workflows, but they remain hand-tuned static mappings: a single 3D lookup table (LUT) or parameter file applied uniformly regardless of image content.
A preset designed for daylight portraiture distorts night-street scenes; one tuned for Kodak Portra~400 behaves unpredictably on images with a radically different dynamic range.

The field's answer to content-blind presets is to make the mapping image-adaptive.
A lightweight network predicts per-image weights over a fixed bank of basis 3D LUTs and blends them~\cite{zeng2022lut,yang2022adaint,yang2022seplut}; because a LUT applies in fixed time regardless of resolution, the design runs in real time, and conditioning the predictor on a reference extends it to several looks~\cite{song2021star}.
The appeal is real; the mechanism goes unexamined.
The blend is a softmax-gated mixture of experts---an ensemble of specialist submodels selected per input by a learned gate---and it inherits the failure that the mixture-of-experts literature documents~\cite{shazeer2017moe}: trained end-to-end against a reconstruction loss, the gate collapses onto one expert.
In a five-LUT model, a single table draws 95\% of the gate weight while the remaining four receive negligible gradient; mean gate entropy is 0.16 of a possible $\ln 5 = 1.61$.
Provisioning $K$ LUTs buys the capacity of roughly one.

Collapse is treatable.
An entropy penalty---the enhancement-setting analogue of mixture-of-experts load balancing---restores near-uniform utilization and recovers about 1~dB PSNR (0.7--1.0~dB at matched training budget; Section~\ref{sec:ablation}).
The treatment then exposes the more basic constraint.
A fixed LUT basis is closed-set: the looks it can produce are fixed when the basis is trained.
For reference-driven film emulation---point the system at any scan and reproduce its look---a closed basis cannot generalize.
And the basis earns little.
An entropy-regularized five-LUT bank reaches 22.16~dB on FiveK Expert~C; a single \emph{conditioned} LUT lands within 0.11~dB on its own self-supervised protocol---not a controlled comparison (Section~\ref{sec:discussion}), but enough to indicate the bank adds little.
The paradigm is at once fragile and bounded, and adding experts is not the lever that matters.

The diagnosis redirects the design.
The obvious alternative---predict one LUT directly---carries a known failure mode: a 33$\times$33$\times$33 LUT has over 100K free parameters, and regressing them without constraint yields the non-smooth, non-monotonic color maps that the basis-and-blend design was built to avoid.
We remove that failure mode rather than route around it.
Parameterizing the predicted LUT as a residual over the identity map holds the output near-identity at initialization and small in magnitude throughout training, which makes direct, single-LUT prediction stable with no basis and no gate (Section~\ref{sec:method}).
We introduce \textit{StyleLUTNet}, a reference-conditioned LUT predictor with no gate to collapse and no basis to constrain it: any reference yields its corresponding LUT (Figure~\ref{fig:teaser}).
Training is self-supervised on procedurally generated color transforms, so no paired film data is required.
A second degeneracy surfaces here: when auxiliary regularization outweighs reconstruction, a learnable residual-scale parameter drifts to zero and the predicted LUT reverts to identity.
The symptom differs from gate collapse, but the cause is the same---the optimizer minimizes a secondary loss at the expense of the primary task.
Both point to one operating principle, which we state and apply: auxiliary LUT regularizers must stay orders of magnitude below the reconstruction loss at convergence.

Color is half of film.
Deep Analog pairs the conditional LUT with a physics-informed optical branch.
Cumulative-distribution-function (CDF) tone matching aligns the graded image to the reference's tonal distribution; a multi-scale grain model varies density with luminance, following the statistics of the emulsion's light-sensitive silver-halide crystals; per-channel halation blurs highlights at wavelength-scaled radii.
We introduce an inverse network, \textit{FilmAnalyzerNet}, that regresses these physical parameters from the reference, closing the loop between reference appearance and rendering.
Decoupling color from texture yields a portable preset---a \texttt{.cube} LUT and a \texttt{.json} parameter file---applied to any image without re-running the network.

This work makes four contributions.
\textbf{(1) A structural diagnosis of the multi-LUT paradigm.}
We identify image-adaptive multi-LUT blending as a softmax-gated mixture and show it inherits expert collapse: under end-to-end training, effective capacity reduces to a single LUT.
We quantify the collapse and show that an entropy term is the decisive intervention---necessary, and with temperature scaling sufficient in our ablation, for a PSNR recovery of about 1~dB.
\textbf{(2) A conditional single-LUT alternative.}
Because a fixed basis is both collapse-prone and closed-set, we replace it with a 3D LUT predicted from a reference (StyleLUTNet), trained by self-supervision; it reaches comparable PSNR on its protocol while generalizing open-set to unseen film stocks.
\textbf{(3) A unifying account of degeneracy in LUT learning.}
A second failure, residual-scale collapse, shares the root cause; we give the operating principle and the loss balance it implies.
\textbf{(4) A physics-informed film renderer with inverse parameter estimation.}
Differentiable grain and halation, driven by FilmAnalyzerNet, with portable \texttt{.cube}/\texttt{.json} preset export.

Three questions organize the investigation.
\textbf{RQ1:} Does the image-adaptive multi-LUT paradigm suffer expert collapse under end-to-end training, and what minimal intervention prevents it?
\textbf{RQ2:} Can a single reference-conditioned LUT match the fixed-basis bank's reconstruction quality while generalizing open-set to unseen film stocks?
\textbf{RQ3:} Can a physics-informed renderer, driven by parameters regressed from one reference, reproduce film grain and halation without per-stock supervision?
The experiments and discussion that follow address them in turn.

\section{Related Work}
\label{sec:related}

\subsection{Image-Adaptive 3D Lookup Tables}

Image-adaptive enhancement predicts a color transform per image rather than applying a fixed one, and the 3D LUT is the transform of choice: a discretized $\R^3\!\to\!\R^3$ map applied by trilinear interpolation at fixed cost, independent of resolution.
Zeng~\etal~\cite{zeng2022lut} established the paradigm---a lightweight CNN predicts per-image weights over a bank of learnable basis LUTs and blends them, reaching sub-millisecond inference after the weight prediction step.
Subsequent work refined the paradigm without altering its structure: AdaInt~\cite{yang2022adaint} adapts the sampling intervals to concentrate LUT precision where perception demands it; SepLUT~\cite{yang2022seplut} factors the 3D LUT into separable 1D components to cut memory.
A neural-implicit variant, NILUT~\cite{conde2024nilut}, replaces the sampled lattice with a coordinate network and blends a fixed set of embedded styles---continuous in color, but still closed over the styles fixed at training.

Across the family, the per-image weights form a softmax gate over a fixed LUT basis---structurally, a mixture of experts.
Reproducing this design, we find the gate collapses onto a single expert under end-to-end training (Section~\ref{sec:swinlut}), an instability analogous to the one load-balancing losses were introduced to counter in large mixture-of-experts models~\cite{shazeer2017moe}.
The enhancement setting has not reported it, and the reason is structural: banks are small ($K\!=\!3$) and evaluation rewards reconstruction alone, so a collapsed gate still scores well.

A second limitation spans the family: 3D LUTs are pixel-independent mappings from $\R^3\!\to\!\R^3$.
Two pixels of identical RGB value receive identical output regardless of spatial context.
This precludes modeling any spatially varying phenomenon---grain, halation, local contrast adaptation---from within the LUT framework alone.
Recent work restores some spatial adaptivity by modulating the LUT through a bilateral grid~\cite{kim2024bglut}, but the objective stays photo enhancement, and film grain and halation remain optical effects outside a color transform, however it is spatially conditioned.

\subsection{Style-Conditioned Enhancement}

Reproducing an arbitrary reference, rather than a fixed retouching target, requires conditioning the transform on that reference.
StarEnhancer~\cite{song2021star} takes the first step, conditioning the weight predictor on an external style reference and enabling multi-style enhancement from a single model; effective for color transfer, it still gates a fixed basis and inherits the pixel-independence limitation.
Broader neural style transfer methods~\cite{gatys2016style,huang2017adain} compute dense feature-space correspondences but in practice yield hallucinated edges at high-contrast boundaries and clipped highlights where the network does not model the gradual shoulder rolloff of film sensitometry.
Classical statistical color transfer~\cite{reinhard2001color}, which aligns the per-channel mean and variance of source and target in a decorrelated color space, avoids these artifacts but cannot capture a film stock's non-linear tonal response.
Reference conditioning is thus established; open-set generalization from a single predicted LUT is not.

\subsection{Learned Color Enhancement and Film Emulation}

Learned enhancement spans several output parameterizations beyond the LUT.
HDRNet~\cite{gharbi2017hdrnet} predicts coefficients of a locally affine transform in a bilateral grid, reaching real-time speed with spatially varying color but bound to a single retouching target learned from paired data.
NeuralPreset~\cite{ke2023neuralpreset} predicts a deterministic color mapping framed explicitly as a ``preset'' and switches styles without re-inference; it shares our goal of an exportable, content-independent mapping, yet learns from a curated style set and targets general color style rather than the tonal and optical signature of a film stock.
Unpaired translation~\cite{zhu2017cyclegan} removes the paired-data requirement through cycle consistency, but trains one generator per domain and yields a network, not a portable preset.
Diffusion-based editors~\cite{brooks2023instructpix2pix} restyle an image from a text or reference instruction at high quality, but inference is iterative rather than real-time and leaves no reusable color transform to export as a preset.
Closest to our setting, Film-GAN~\cite{gong2024filmgan} translates digital photographs to an analog-film look with a GAN, separating a reference's color and grain much as we do; but as a generator it returns a network rather than an editor-native preset and carries the training instability of adversarial objectives, whereas we predict a deterministic, exportable color transform paired with an explicit optical model.
Deep Analog differs on three axes: the color mapping is a 3D LUT conditioned on an arbitrary reference (open-set), trained without paired data, and exported as a \texttt{.cube} file; and color is paired with an explicit physical model of grain and halation rather than left to a black-box generator.

\subsection{Vision Transformers for Low-Level Vision}

Swin Transformer~\cite{liu2021swin} introduced hierarchical shifted-window self-attention that scales linearly with image resolution.
Uformer~\cite{wang2021uformer} demonstrated Transformer encoders outperform CNN counterparts on image restoration, attributing gains to non-local feature aggregation.
Chen~\etal~\cite{chen2021ipt} showed pre-trained Transformer backbones transfer effectively to multiple image processing tasks.
These backbones motivate our weight predictor: we instantiate the multi-LUT paradigm with a Swin-T encoder (Section~\ref{sec:swinlut}) to probe it at full strength, so that the collapse we report reflects the paradigm rather than an underpowered backbone.

\subsection{Film Physics Simulation}

The spatially varying phenomena a LUT cannot express---grain, halation---have established physical models outside the enhancement literature.
Ameur~\etal~\cite{ameur2023grain} developed a deep learning approach to film grain synthesis modeling intensity-dependent noise statistics of silver halide emulsion.
Spencer~\etal~\cite{spencer1995glare} established the physical model for optical glare, defining point-spread functions for light scatter in emulsion layers.
Wu~\etal~\cite{wu2021flare} formulated a physics-based synthesis pipeline for lens flare, demonstrating that physically generated training data can supervise a neural network to reproduce optical artifacts.
These models render a phenomenon from known parameters; none estimates those parameters from a reference scan, which is what film emulation from an arbitrary target demands.

\subsection{Histogram Specification and Tone Transfer}

Histogram alignment underlies our tone matching and predates learned methods.
Exact histogram specification~\cite{coltuc2006exact} imposes a target histogram by strictly ordering pixels; the ordering is discrete, so when source and reference dynamic ranges differ, the inverse mapping produces visible banding.
Reinhard~\etal's statistical transfer~\cite{reinhard2001color} avoids banding by matching only per-channel mean and variance in a decorrelated space, at the cost of discarding higher moments of the distribution.
Our CDF-based tone curve matching (Section~\ref{sec:tone_curve}) takes the middle ground: it aligns the full per-channel distribution like exact specification, but smooths the transfer function with a Gaussian kernel and adds triangular dithering to suppress that banding.
Reinhard~\etal's tone-reproduction operator~\cite{reinhard2002tone} separately motivates the smooth highlight shoulder we reproduce in film tone processing (Section~\ref{sec:film_tone}).

\subsection{Self-Supervised Learning}

Self-supervised learning derives a training signal from unlabeled data through a pretext task.
Contrastive methods~\cite{chen2020simclr} learn representations by pulling together augmented views of one image and pushing apart views of others; the augmentations---color jitter, blur, crop---supply the supervision.
StyleLUTNet adapts the principle to a different end.
Rather than learn an invariant representation, it learns to \emph{invert} a known color augmentation: given a reference and a target produced by the same random transform, the network must recover the transform's effect as a 3D LUT.
The supervision is exact---the target pixels are known---rather than contrastive, which removes negative sampling and lets the loss act directly in pixel and histogram space.

\subsection{Positioning}

Three gaps run through this literature. The multi-LUT family is collapse-prone under end-to-end training and closed-set, its looks fixed once the basis is trained. Reference-conditioned methods open the look but still gate a fixed basis and treat color pixel-independently. Physical grain and halation models render from known parameters but do not estimate them from a reference. Deep Analog addresses the three together: a single 3D LUT predicted from an arbitrary reference---no gate, no basis---trained by self-supervision, paired with a physical optical model whose parameters are regressed from the same reference.

\section{Method}
\label{sec:method}

\begin{figure*}[t]
  \centering
  \includegraphics[width=\textwidth]{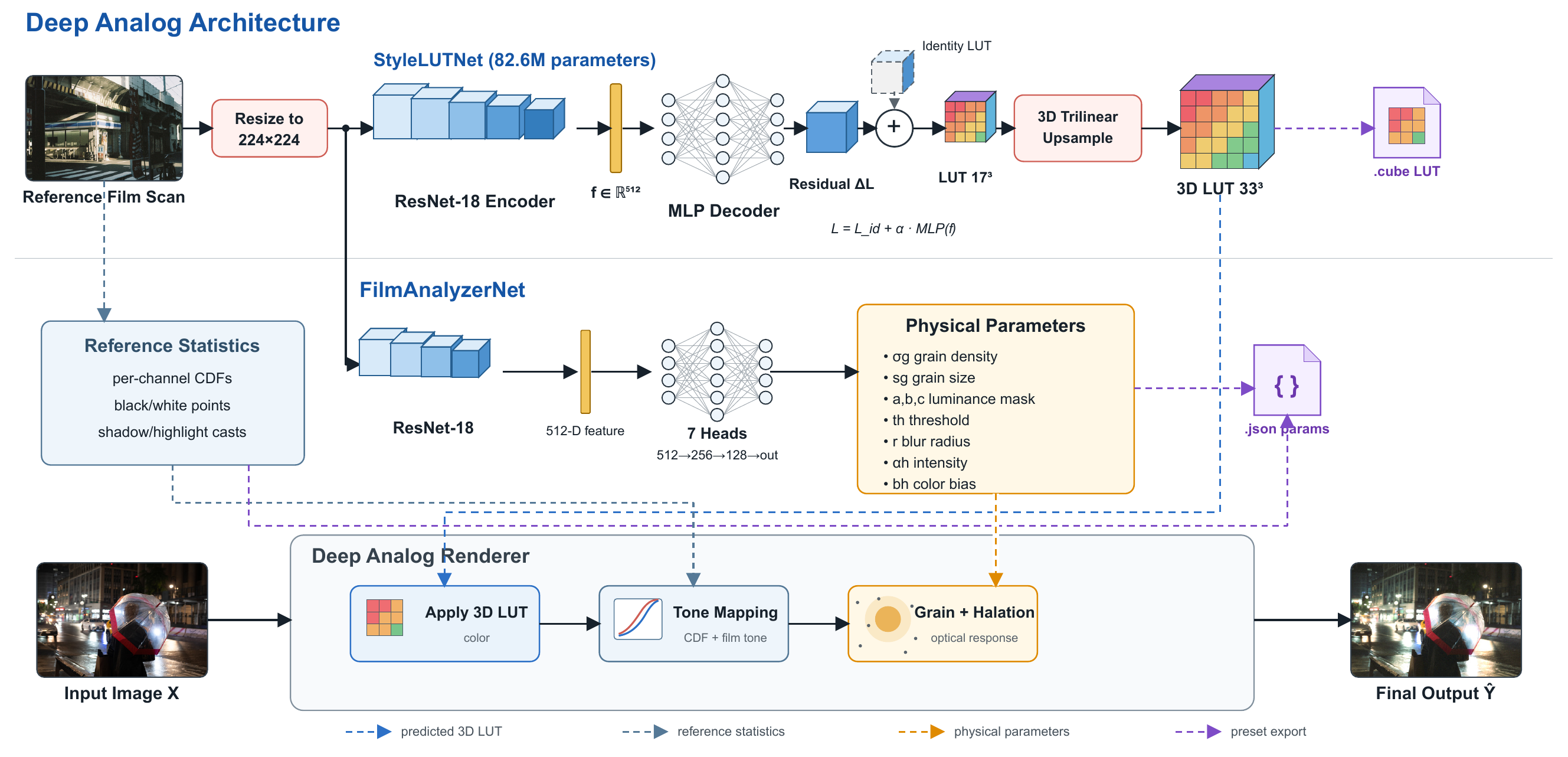}
  \caption{Deep Analog architecture. StyleLUTNet encodes a resized reference scan into a residual $17^3$ LUT, adds it to a fixed identity LUT, and upsamples the result to $33^3$. In parallel, FilmAnalyzerNet predicts seven groups of physical parameters (11 scalar outputs), while a deterministic analyzer measures reference CDFs, black and white points, and shadow/highlight color casts. The predicted LUT, reference statistics, and physical parameters condition the color, tone, and optical components of the renderer, respectively. The system exports the LUT as a \texttt{.cube} file and combines tone statistics with physical rendering parameters in a \texttt{.json} preset.}
  \label{fig:architecture}
\end{figure*}

\subsection{System Overview}

Deep Analog processes a target photograph conditioned on a reference film scan through a six-stage pipeline:

\begin{enumerate}[leftmargin=*,itemsep=2pt]
    \item \textbf{Reference analysis.} FilmAnalyzerNet extracts grain and halation parameters from the reference; a statistical analyzer measures tonal characteristics (black point, white point, shadow and highlight color casts).
    \item \textbf{StyleLUT color grading.} StyleLUTNet predicts a conditional 3D LUT from the reference and applies it to the full-resolution target via trilinear interpolation.
    \item \textbf{Tone curve matching.} Per-channel CDF-based histogram transfer aligns the graded image's tonal distribution to the reference.
    \item \textbf{Film tone processing.} Analytic adjustments---black point lift, highlight rolloff, shadow and highlight color tinting---reproduce tonal characteristics that the neural LUT alone cannot capture.
    \item \textbf{Multi-scale grain synthesis.} Three-octave noise, weighted and modulated by a luminance-dependent mask, adds film grain at physically motivated scales.
    \item \textbf{Per-channel halation.} Highlight regions are extracted, blurred at wavelength-dependent radii, and composited with learned color bias.
\end{enumerate}

\begin{figure*}[t]
  \centering
  \includegraphics[width=\textwidth]{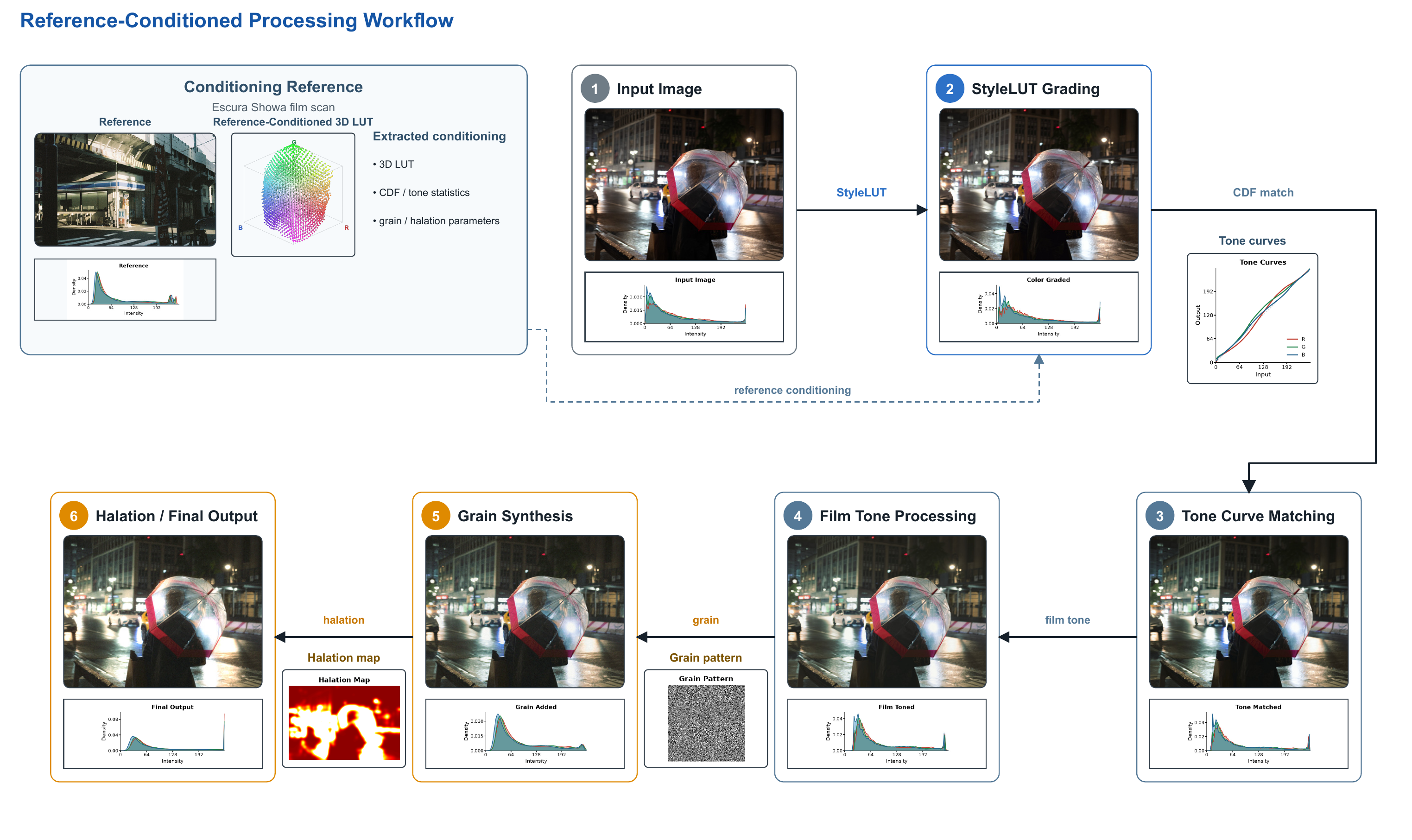}
  \caption{Stagewise processing workflow for Rainy Night NYC conditioned on an Escura Showa film scan. The reference card includes the predicted 3D LUT, whose point positions and colors encode output RGB coordinates. The numbered sequence follows a two-row path: input, StyleLUT grading, CDF tone-curve matching, analytic film-tone processing, grain synthesis, and halation rendering. Each card pairs the image with its RGB histogram; diagnostic panels beneath the corresponding transition arrows visualize the CDF-derived tone-transfer curve, grain residual, and halation map. Grain and halation use illustrative amplified parameters for print visibility.}
  \label{fig:workflow}
\end{figure*}

Stages 2--4 handle the \textit{photometric response} (color and tone); stages 5--6 handle the \textit{optical response} (grain and halation).
The predicted LUT from stage~2 exports to the \texttt{.cube} format, while the reference-derived tone-transfer curves and physical parameters export to JSON; together they form a portable preset.
Figure~\ref{fig:architecture} summarizes the model and conditioning paths; Figure~\ref{fig:workflow} shows the visible image progression through the six stages.

\subsection{StyleLUTNet: Conditional 3D LUT Prediction}
\label{sec:stylelut}

The color transform demands learning; the analytic tone steps that follow---CDF matching and film-tone processing (Sections~\ref{sec:tone_curve},~\ref{sec:film_tone})---do not.
Those steps read simple aggregate statistics---percentile luminances, mean colors of tonal regions---from pixels present in the reference, and any reference determines them in closed form.
A 3D LUT is different in kind: its domain is the whole RGB cube, so it must return an output for nearly every input color.
That includes colors absent from the given reference---a night-street reference may contain no warm skin tones, yet must still grade a portrait target.
No statistic over the reference's own pixels can answer where the reference has no data.
Only a network trained across many known color transforms (the self-supervised RandomColorTransform protocol, Section~\ref{sec:stylelut_training}) acquires a general rule for color transformation that extrapolates to such unseen colors; a closed-form extraction cannot.

StyleLUTNet predicts a per-reference 3D lookup table through residual learning.
A ResNet-18 encoder~\cite{he2016resnet}, pre-trained on ImageNet-1K, processes the reference image at $224\!\times\!224$---its ImageNet pretraining resolution---and produces a 512-dimensional feature vector $\mathbf{f} \in \R^{512}$.
A light encoder suffices here: the target is a global color operator, so the encoder's task is to summarize the reference's global color statistics rather than model long-range spatial structure---the regime where attention earns its cost.
The capacity we grant the Swin-T probe (Section~\ref{sec:swinlut}) is not needed, and parameters are better spent in the decoder that carries the LUT output (71.4M of the model's 82.6M).
An MLP decoder maps $\mathbf{f}$ to a residual LUT:
\begin{equation}
    \Phi_{\text{pred}} = \Phi_{\text{id}} + \alpha \cdot \text{MLP}(\mathbf{f}),
    \label{eq:stylelut}
\end{equation}
where $\Phi_{\text{id}} \in \R^{3 \times D_l \times D_l \times D_l}$ is a fixed identity LUT (mapping each RGB value to itself), $\alpha$ is a fixed scale factor, and $D_l = 17$ is the low-resolution grid dimension.
The decoder is a four-layer MLP with widths 1024, 2048, and 4096, widening toward the high-dimensional LUT output; LayerNorm, GELU, and dropout ($p\!=\!0.1$) follow the standard defaults for MLP decoders rather than any task-specific tuning.
The final layer is zero-initialized, so the network starts from the identity mapping.

The decoder predicts the LUT at $17^3$ rather than $33^3$, which reduces its output dimensionality from $3 \times 33^3 = 107{,}811$ to $3 \times 17^3 = 14{,}739$ (a 7.3$\times$ reduction).
This is the sole purpose of the low-resolution grid: fewer decoder outputs mean fewer parameters and a more stable regression target; it does not trade away color accuracy, because the LUT is a smooth transform that a $17^3$ lattice already samples adequately.
The predicted $17^3$ LUT is upsampled to $33^3$ by 3D trilinear interpolation and applied to the full-resolution target with \texttt{F.grid\_sample}.
The upsampling adds no color information---trilinear interpolation over the $17^3$ lattice is, up to negligible resampling error, the same transform that applying the $17^3$ LUT directly would produce.
Its only role is interface: $33^3$ is the grid size of the baselines we compare against~\cite{zeng2022lut,yang2022adaint} and the common resolution for exported \texttt{.cube} files, so predicting at $17^3$ and emitting at $33^3$ keeps the model lightweight while matching the standard evaluation and deployment format.

Residual prediction is what makes the direct route stable.
The basis-and-blend design avoided regressing a full LUT by restricting its output to a low-dimensional mixture of a few jointly learned, already well-behaved tables---stability by construction, at the price of a fixed basis.
The residual parameterization achieves the same stability without the basis: zero initialization starts every prediction at the identity operator, the network learns only a deviation from it, and the regularizers of Section~\ref{sec:losses} keep that deviation smooth and monotone.

\paragraph{Residual scale.}
Early experiments used a learnable scalar $\alpha$ initialized to 0.1 as an \texttt{nn.Parameter}.
When smoothness and monotonicity regularization weights were set too high relative to the reconstruction loss (a ratio of 15:1 in our case), the optimizer found that shrinking $\alpha$ toward zero reduced the regularization penalty more efficiently than improving color accuracy.
This caused $\alpha$ to collapse from 0.1 to 0.04 over 100 training epochs, producing near-identity output regardless of the reference.
Fixing $\alpha = 1.0$ as a non-learnable buffer and rebalancing loss weights (Section~\ref{sec:stylelut_training}) resolved the issue.

\paragraph{Self-supervised training.}
StyleLUTNet requires no paired film-scan datasets.
Training uses a \textit{RandomColorTransform} procedure: given two images $I_a$ and $I_b$, a random composition of 2--4 color transformations (gamma curves in [0.3, 3.0], color temperature shifts of $\pm$0.30, HSV adjustments with saturation in [0.25, 2.0], split toning, $3\!\times\!3$ color matrix perturbation with noise scale 0.20, and levels adjustments) is applied identically to both.
The ranges are set to span, and slightly exceed, the color variation across real film stocks, so the learned inverse extends to looks outside the training distribution.
Image $I_a$ serves as the reference; the transformed $I_b$ serves as the target; the original $I_b$ is the ground truth.
This forces the network to learn the inverse of arbitrary color transformations, generalizing to film looks it has never seen.

\subsection{SwinLUT: Multi-LUT Architecture and Weight Collapse}
\label{sec:swinlut}

To study the paradigm in its own terms, we build a model in its image: a bank of $K\!=\!5$ learnable basis LUTs blended by softmax weights, with a Swin-T~\cite{liu2021swin} encoder predicting the weights in place of the CNN of~\cite{zeng2022lut}.
We set $K\!=\!5$, larger than the $K\!=\!3$ of prior work, so the bank has room to spread its weight and any collapse onto a single expert is unambiguous rather than an artifact of a small basis.
The Transformer produces $\mathbf{v}_{\text{style}} \in \R^{768}$, passed through an MLP to predict the softmax weights:
\begin{equation}
    \mathbf{w} = \sigma\!\left(\frac{\text{MLP}(\mathbf{v}_{\text{style}})}{\tau}\right), \quad
    \Phi_{\text{fused}} = \sum_{k=1}^{K} w_k \, \Phi_k,
\end{equation}
where $\tau = \exp(\theta_\tau)$ is a learnable temperature clamped to $[0.1, 10.0]$; larger $\tau$ flattens the predicted weight distribution, smaller $\tau$ sharpens it.

\paragraph{Weight collapse.}
Training the baseline (standard softmax, no regularization) collapsed the gate onto a single expert: one basis LUT dominated while the other four received negligible gradient and were effectively unused.
We quantify the collapse---the dominant weight and the gate entropy relative to its $\ln K$ maximum---in Section~\ref{sec:swinlut_dev}.
This is a predictable consequence of softmax-gated mixture models: slight early advantages compound through a positive feedback loop where the dominant expert receives more gradient, becomes more accurate, and attracts even higher weight.
The mechanism is a property of the gate, not the encoder: it depends only on softmax dynamics under a reconstruction objective, so the choice of backbone---Swin-T here, a plain ViT, or a CNN---does not avert the collapse.
We adopt Swin-T precisely to instantiate the paradigm with a strong modern backbone, ruling out the objection that the collapse reflects an underpowered predictor rather than the paradigm itself.

\paragraph{Mitigation.}
Three interventions restore weight diversity: (1)~temperature scaling, where $\tau$ converges to $\approx$2.2, broadening the softmax distribution; (2)~entropy regularization $\mathcal{L}_{\text{ent}} = \sum_k w_k \log w_k$, which penalizes peaked distributions; and (3)~pairwise diversity loss $\mathcal{L}_{\text{div}} = \frac{1}{\binom{K}{2}} \sum_{i<j} |\cos(\text{vec}(\Phi_i), \text{vec}(\Phi_j))|$, which prevents basis LUTs from converging to identical mappings.
A full ablation in Section~\ref{sec:ablation} confirms that entropy regularization alone accounts for the majority of the recovery, while diversity loss is marginal when entropy regularization is active.

Collapse mitigation makes the bank usable but cannot lift its ceiling: a fixed basis still requires paired training data and still defines a closed set of looks.
That closed-set limitation, not any defect of the architecture, motivates the conditional StyleLUTNet design (Section~\ref{sec:stylelut}).

\subsection{Film Physics Module}
\label{sec:physics}

This module reproduces the film characteristics that the color LUT cannot, in two groups.
The photometric response---tonal distribution and the tone-curve properties of a film stock---is handled by CDF-based tone matching (Section~\ref{sec:tone_curve}) and analytic film-tone processing (Section~\ref{sec:film_tone}).
Unlike the LUT of Section~\ref{sec:stylelut}, these tonal steps need no learning: each is a closed-form aggregate over pixels already present in the reference---a percentile, a regional mean color---so any reference determines it directly, with no need to answer for colors the reference never contains.
The optical response comprises two spatially varying phenomena absent from any pixel-independent LUT---grain and halation (Sections~\ref{sec:grain},~\ref{sec:halation}).
We model each through differentiable rendering functions parameterized by values extracted from the reference scan, grounded where possible in the sensitometry---the quantitative relation between exposure and developed density---and physics of the photographic process~\cite{james1977photographic}.

\subsubsection{Tone Curve Matching}
\label{sec:tone_curve}

Before applying grain and halation, we align the tonal distribution of the LUT-graded image to the reference via per-channel CDF matching.
For each color channel, we construct cumulative distribution functions at 256 uniform sample points, matching the 8-bit quantization grid:
\begin{equation}
    F_{\text{src}}(v) = \frac{1}{N}\sum_{i=1}^{N} \mathbf{1}[x_i \leq v], \quad
    F_{\text{ref}}(v) = \frac{1}{M}\sum_{j=1}^{M} \mathbf{1}[y_j \leq v],
\end{equation}
where $x_i$ and $y_j$ are pixel values in source and reference, respectively.
The transfer function $T(v) = F_{\text{ref}}^{-1}(F_{\text{src}}(v))$ is computed via binary search (\texttt{torch.searchsorted}) and smoothed with a 1D Gaussian kernel ($\sigma\!=\!2$ bins, kernel size 9) to eliminate quantization steps.
Application uses linear interpolation between adjacent bin centers.

Naive sort-based histogram matching, in the manner of exact histogram specification~\cite{coltuc2006exact}, produces severe banding when the source and reference have very different dynamic ranges---for example, transferring a dark indoor scan to a bright outdoor target; matching only per-channel mean and variance~\cite{reinhard2001color} avoids banding but discards the higher moments of the tonal distribution.
Our CDF-based approach with Gaussian smoothing yields a continuous transfer function that aligns the full distribution while suppressing these artifacts.
A final triangular-distribution dithering step at $\pm 0.5/256$ amplitude breaks any residual quantization artifacts.

\subsubsection{Film Tone Processing}
\label{sec:film_tone}

The neural LUT captures global color shifts but misses several tonal properties intrinsic to analog film.
We address these through analytic adjustments derived from reference statistics:

\paragraph{Reference analysis.}
From the reference scan, we measure: black point (2nd percentile luminance), white point (98th percentile), mean color in the darkest 15\% of pixels (shadow color cast), and mean color in the brightest 15\% (highlight color cast).
The percentile endpoints and the 15\% region size are hand-set---extreme enough to stay in the tonal tails, wide enough to average scene texture out of the cast estimate.

\paragraph{Black point lift.}
Analog film does not clip to pure black; unexposed regions retain a minimum density ($D_{\min}$) set by the film base plus fog level, the toe of the characteristic curve---the S-shaped density-versus-log-exposure response of a stock~\cite{james1977photographic}.
When the reference black point exceeds 0.02, we lift the image shadows:
\begin{equation}
    I' = I \cdot (1 - \beta) + \beta, \quad \beta = b_{\text{ref}} \cdot s \cdot 0.8,
\end{equation}
where $b_{\text{ref}}$ is the measured reference black point and $s$ is a user-controllable strength parameter.
The affine form is our own: a minimal, monotone lift driven directly by the measured $D_{\min}$ rather than a fitted density curve.
The 0.02 activation threshold and 0.8 damping factor are hand-set to keep the lift conservative: it engages only when the reference demonstrably lifts blacks, and never transfers the full measured amount.

\begin{figure}[tbp]
  \centering
  \includegraphics[width=\linewidth]{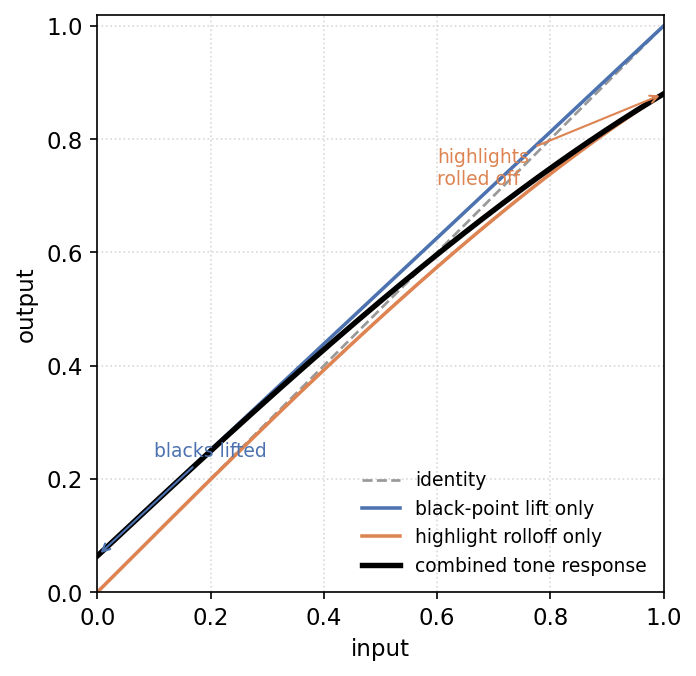}
  \caption{Analytic tone response (Section~\ref{sec:film_tone}). The black-point lift (blue) raises the toe; the quadratic highlight rolloff (orange) softens the shoulder; together (black) they compress the tonal range into the lifted-black, rolled-highlight response characteristic of scanned film, using two parameters ($b_{\text{ref}}$, $w_{\text{ref}}$) read from the reference. This step is applied after the neural LUT and CDF matching, which carry the bulk of the tonal transfer, so it remains a small residual correction---unlike the display-referred operators (Reinhard~\cite{reinhard2002tone}, Naka--Rushton~\cite{naka1966spotentials}, ACES filmic) that a full sigmoid fit would require.}
  \label{fig:tone_curves}
\end{figure}

\paragraph{Highlight rolloff.}
Film exhibits a smooth shoulder where high-exposure values compress gradually rather than clipping hard---the upper bend of the characteristic curve~\cite{james1977photographic}, the analogue of which motivates photographic tone-reproduction operators~\cite{reinhard2002tone}.
We model this with a quadratic rolloff:
\begin{equation}
    I' = I \cdot \left(1 - I^2 \cdot (1 - c)\right), \quad c = 1 - (1 - w_{\text{ref}}) \cdot s,
\end{equation}
where $w_{\text{ref}}$ is the reference white point.
The quadratic term $I^2$ ensures the rolloff is smooth with no discontinuity at the transition point---unlike a piecewise linear knee, which introduces visible banding at the boundary.
This single-parameter quadratic is our own differentiable stand-in for the characteristic-curve shoulder, trading sensitometric exactness for smoothness and training stability.
Standard S-curve operators---the Reinhard photographic operator~\cite{reinhard2002tone}, the Naka--Rushton saturating sigmoid~\cite{naka1966spotentials}, and the ACES filmic curve used in film production---model toe and shoulder jointly with a more principled response.
We do not adopt one because the analytic tone step is a residual correction: the neural LUT and CDF matching (Section~\ref{sec:tone_curve}) already carry the bulk of the tonal transfer, so the remaining shoulder needs only a monotone, single-parameter, differentiable approximation, not a full sensitometric fit.
Figure~\ref{fig:tone_curves} shows the resulting transfer: the black lift and quadratic rolloff compress the tonal range into the lifted-black, rolled-highlight response characteristic of scanned film, from two parameters derived from reference statistics.

\paragraph{Shadow and highlight tinting.}
Film stocks carry characteristic color casts in shadows and highlights (Fuji Superia's green shadows, Kodak Portra's warm highlights).
Let $\mathbf{c}_{\text{shadow}} \in \R^3$ be the mean RGB of the darkest 15\% of reference pixels (Reference analysis, above), and $\bar{c}_{\text{shadow}} = \tfrac{1}{3}\sum_{k} c_{\text{shadow},k}$ its mean across channels---the neutral-gray level of that region.
Subtracting the two yields the region's \emph{color cast}, its chromatic deviation from neutral:
\begin{equation}
    \mathbf{t}_{\text{shadow}} = \mathbf{c}_{\text{shadow}} - \bar{c}_{\text{shadow}}\,\mathbf{1},
    \qquad \mathbf{1}=(1,1,1)^\top,
\end{equation}
and $\mathbf{t}_{\text{highlight}} \in \R^3$ is defined identically from the brightest 15\%.
We transfer both casts to the target image $I$, each confined to its own tonal range by a luminance mask:
\begin{equation}
    I' = I + (1 - L)^2\,\mathbf{t}_{\text{shadow}} + L^2\,\mathbf{t}_{\text{highlight}},
    \label{eq:tinting}
\end{equation}
where $L\in[0,1]$ is per-pixel luminance; the quadratic exponents concentrate each tint in the tonal extremes and leave mid-tones near-neutral.
The construction is our own: a tonally localized form of split toning, related to statistical color transfer~\cite{reinhard2001color} but restricted to the zero-sum chroma residual $\mathbf{t}$ of the reference's tonal extremes, so it shifts color balance while leaving the mean channel level unchanged.

\subsubsection{Multi-Scale Grain Synthesis}
\label{sec:grain}

Real film grain spans a range of spatial frequencies, arising from the intensity-dependent statistics of developed silver halide crystals~\cite{james1977photographic,ameur2023grain}.
Fine grain from individual crystals sits at single-pixel scale; clumping produces medium-scale structure; scanner-induced correlation adds coarser texture.
Following the stochastic, luminance-dependent modeling of film grain~\cite{newson2017filmgrain}, we synthesize grain at three octaves with relative weights 50\%, 33\%, and 17\%, decreasing with scale so the finest grain dominates, consistent with the frequency content of scanned emulsion; the specific octave weighting and per-octave normalization below are our differentiable parameterization:

\begin{equation}
    G = \sum_{o=1}^{3} w_o \cdot \frac{\text{Blur}(\mathbf{n}, \sigma_o)}{\text{std}(\text{Blur}(\mathbf{n}, \sigma_o))},
    \label{eq:grain}
\end{equation}
where $\mathbf{n}$ is per-channel noise composed half of a shared luminance component and half of independent chrominance---a hand-set split that renders grain correlated across color layers but not identical---and $\sigma_o$ increases with octave index.
The blur radii scale with image resolution via a factor $\sqrt{HW}/480$, anchored at the 480-pixel training crop size, to maintain consistent perceptual grain size across output resolutions.

Grain intensity varies with exposure: in real film, shadows are grainier than highlights because fewer silver halide crystals are activated.
We capture this with a learnable luminance mask of our own design---a sigmoid over a quadratic, bounded in $[0,1]$ and smooth, its shape recoverable by regression:
\begin{equation}
    M(L) = \sigma(aL^2 + bL + c),
\end{equation}
where $a$, $b$, $c$ are parameters predicted by FilmAnalyzerNet (Section~\ref{sec:filmanalyzer}).
The final grain is $I' = I + \sigma_g \cdot M(L) \cdot G$, where $\sigma_g$ is the predicted grain intensity.

\subsubsection{Per-Channel Halation}
\label{sec:halation}

Halation occurs when bright light passes through the emulsion, reflects off the film base, and re-exposes the emulsion from behind, producing a soft glow around highlights~\cite{james1977photographic,spencer1995glare}.
Its characteristic warm color is not a wavelength-dependent scattering effect---by Rayleigh scattering shorter wavelengths would spread more, not less.
It follows instead from emulsion structure: the red-sensitive layer lies deepest, nearest the base, so long-wavelength light both penetrates furthest and contributes most to the base-reflected re-exposure, biasing the halo toward red~\cite{james1977photographic}.

Our renderer extracts highlights via a soft threshold:
\begin{equation}
    H = I \cdot \sigma(20 \cdot (L - t_h)),
\end{equation}
where $t_h$ is the halation threshold and the steep sigmoid (slope 20) approximates a step function while remaining differentiable.
The soft threshold and the additive channel-blur compositing that follows are our own differentiable rendering of the effect, not a fitted optical point-spread model~\cite{spencer1995glare}.
Each color channel is blurred independently with per-channel radii: $r_R = 1.4r$, $r_G = 1.0r$, $r_B = 0.7r$, where $r$ is the base halation radius.
The ratio 1.4:1.0:0.7 is set empirically to reproduce the red-biased halo, assigning the widest support to the red channel that dominates the base-reflected re-exposure.
The blurred highlights are weighted by a learned color bias $\mathbf{b}_h \in \R^3$ (passed through sigmoid) and intensity $\alpha_h$, then added to the image:
\begin{equation}
    I' = I + \alpha_h \cdot \sigma(\mathbf{b}_h) \cdot \text{ChannelBlur}(H).
\end{equation}

\begin{figure*}[t]
  \centering
  \includegraphics[width=\textwidth]{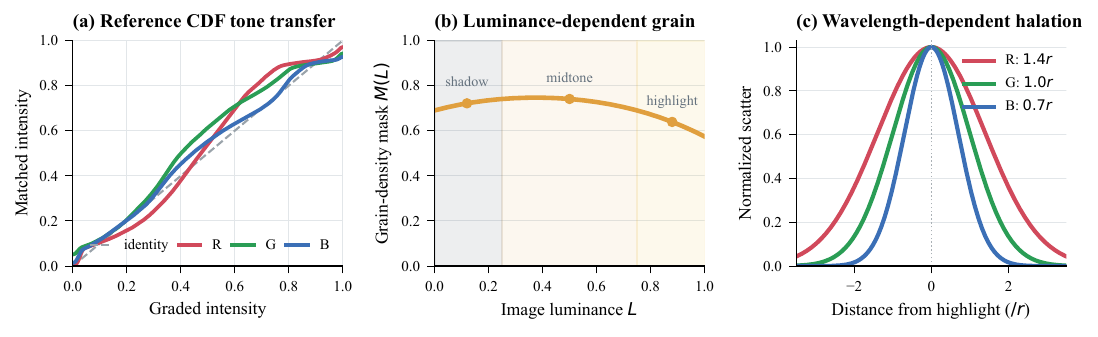}
  \caption{Response functions in the physics-informed renderer. (a)~Per-channel CDF transfer curves computed for the Rainy Night NYC example after StyleLUT grading, conditioned on the Escura Showa reference; the dashed diagonal is the identity mapping. (b)~The luminance-dependent grain mask $M(L)=\sigma(aL^2+bL+c)$ for $a=-2.0$, $b=1.5$, and $c=0.8$, showing strongest response in the mid-tones. (c)~Normalized radial profiles of the per-channel halation kernels. The red kernel has the broadest support ($1.4r$), followed by green ($r$) and blue ($0.7r$), producing the characteristic warm halo around bright regions.}
  \label{fig:physics_responses}
\end{figure*}

\begin{figure*}[t]
  \centering
  \includegraphics[width=\textwidth]{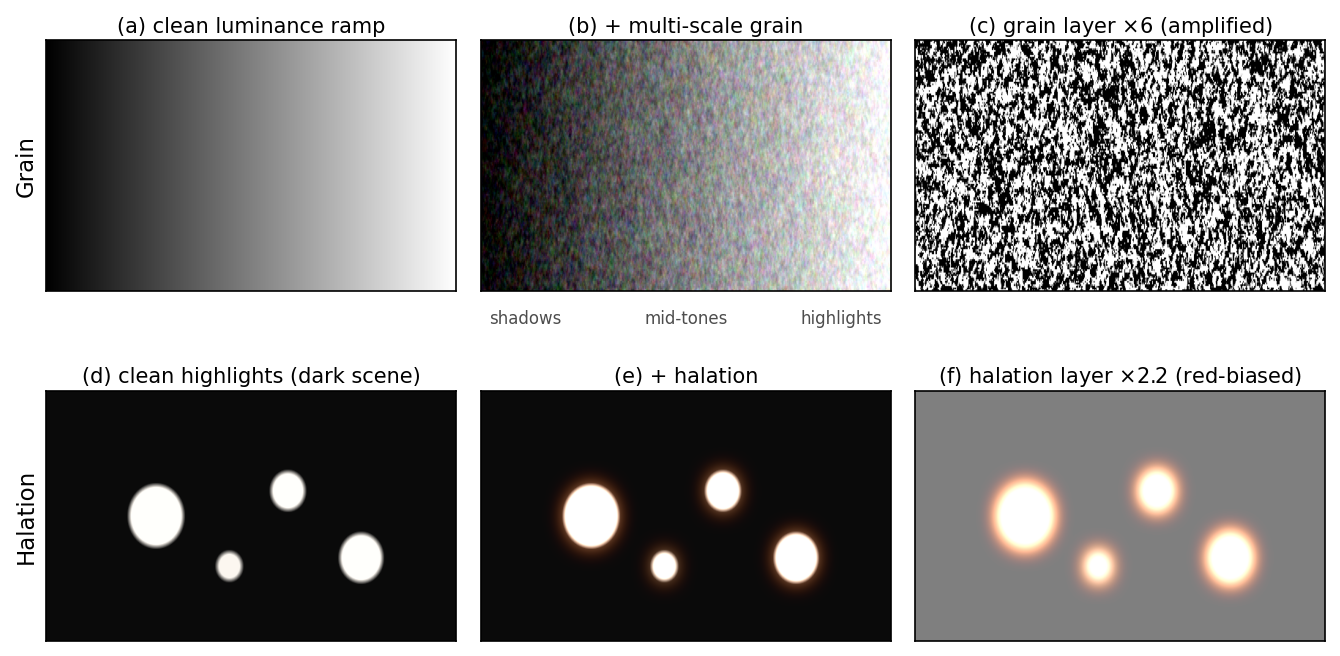}
  \caption{Controlled demonstration of the optical renderer on synthetic test scenes, applying the grain (Section~\ref{sec:grain}) and halation (Section~\ref{sec:halation}) equations directly. These illustrate the rendering functions in isolation; they are not results on real film. \textbf{Top row (grain):} (a)~a clean luminance ramp; (b)~after multi-scale grain, whose amplitude is modulated by the luminance mask $M(L)$; (c)~the grain layer amplified $6\times$. \textbf{Bottom row (halation):} (d)~bright discs on a dark background; (e)~after halation, the highlights acquire a warm glow that spreads beyond their borders; (f)~the halation layer amplified, showing the red-biased spread produced by the per-channel radii $r_R\!:\!r_G\!:\!r_B = 1.4\!:\!1.0\!:\!0.7$.}
  \label{fig:film_texture}
\end{figure*}

Figure~\ref{fig:physics_responses} visualizes the three response functions that are otherwise easy to obscure in the equations.
The tone stage learns a separate monotone mapping for each channel, the grain mask changes continuously with exposure, and the halation kernels encode the per-channel spatial spread.
Only the first is a global color transform; grain and halation remain spatially varying effects that cannot be represented by a 3D LUT alone.
Figure~\ref{fig:film_texture} isolates the two optical effects on controlled synthetic scenes, where each is unambiguous; Figure~\ref{fig:grain_halation} then shows the same effects on real image crops, comparing regions before and after each stage.

\subsection{FilmAnalyzerNet: Inverse Parameter Estimation}
\label{sec:filmanalyzer}


FilmAnalyzerNet takes a $224\!\times\!224$ resized reference image and uses seven heads to regress 11 scalar outputs grouped into the parameters that drive the grain and halation renderers.
The architecture consists of a ResNet-18 backbone producing a 512-dimensional feature vector, followed by seven independent regression heads.
Each head is a three-layer MLP ($512\!\to\!256\!\to\!128\!\to\!\text{output}$) with ReLU activations and dropout, set here to $p\!=\!0.3$---higher than the StyleLUTNet decoder's $0.1$---to counter overfitting on the low-dimensional regression targets.

Output activations enforce physical constraints:
\begin{itemize}[leftmargin=*,itemsep=1pt]
    \item \textbf{Grain intensity} ($\sigma_g$): $\text{softplus} \times 0.05 + 0.005$, ensuring positivity with a minimum floor.
    \item \textbf{Grain size}: $\text{softplus} \times 1.5 + 0.5$, bounding scale to $[0.5, \infty)$.
    \item \textbf{Luminance curve} ($a, b, c$): unconstrained (sigmoid in Eq.~\ref{eq:grain} bounds the mask).
    \item \textbf{Halation threshold} ($t_h$): $\text{sigmoid} \times 0.45 + 0.5$, confining to $[0.5, 0.95]$ (only bright regions).
    \item \textbf{Halation radius} ($r$): $\text{softplus} \times 4.0 + 2.0$, ensuring minimum blur extent.
    \item \textbf{Halation intensity} ($\alpha_h$): $\text{sigmoid} \times 0.5$, capping at moderate strength.
    \item \textbf{Color bias} ($\mathbf{b}_h$): unconstrained (sigmoid applied during rendering).
\end{itemize}

Training uses synthetic film-look data generated by applying the grain and halation renderers with known parameters to clean images, then supervising FilmAnalyzerNet to recover those parameters.
This self-supervised loop avoids the need for ground-truth physical measurements of real film stocks.

\subsection{Loss Functions}
\label{sec:losses}

The three networks train under different objectives, but the objectives share a lineage.
SwinLUT, the diagnostic probe, adopts the reconstruction and LUT-regularity terms of Zeng~\etal~\cite{zeng2022lut} and adds two gate-level regularizers against the weight collapse of Section~\ref{sec:swinlut}.
StyleLUTNet retains the LUT-regularity terms---the pathologies they prevent are properties of the table, not of the gate---drops the gate terms, since a single conditional LUT has no gate to collapse, and adds a histogram term demanded by the self-supervised transfer objective of Section~\ref{sec:stylelut}.
FilmAnalyzerNet is supervised regression and needs only a weighted parameter loss.
Coefficients throughout are governed by the operating principle of Section~\ref{sec:stylelut}: the quantity to control is each weighted term at convergence, which must remain orders of magnitude below reconstruction.
Coefficient symbols are local to each objective; the two LUT losses normalize their regularizers differently, so values are not comparable across subsections.

\subsubsection{StyleLUTNet Training Loss}

StyleLUTNet must reproduce the reference's color distribution while keeping the predicted table a well-behaved color operator:
\begin{equation}
    \mathcal{L}_{\text{style}} = \mathcal{L}_{\text{pixel}} + \lambda_s \mathcal{L}_{\text{smooth}} + \lambda_m \mathcal{L}_{\text{mono}} + \lambda_h \mathcal{L}_{\text{hist}},
    \label{eq:style_loss}
\end{equation}
where $\mathcal{L}_{\text{pixel}}$ is L1 reconstruction loss and each remaining term targets a specific failure mode.
$\mathcal{L}_{\text{smooth}}$, total variation over the predicted LUT, suppresses the oscillatory mappings a directly regressed $33^3$ table can express (Section~\ref{sec:stylelut}); $\mathcal{L}_{\text{mono}}$ penalizes monotonicity violations, which surface as tone reversal; $\mathcal{L}_{\text{hist}}$, a per-channel histogram matching loss, supervises the actual objective---the self-supervised task is distribution transfer, which per-pixel error alone underconstrains.
Coefficients $\lambda_s = 1.0$, $\lambda_m = 0.5$, $\lambda_h = 3.0$ were hand-tuned on the validation split; the histogram term receives the largest weight for the reason above.

\paragraph{Regularization balance.}
The specific values matter less than the constraint they satisfy.
Setting $\lambda_s$ and $\lambda_m$ too high relative to $\mathcal{L}_{\text{pixel}}$ causes the residual scale collapse described in Section~\ref{sec:stylelut}; at the adopted values, the weighted smoothness and monotonicity terms converge to ${\sim}10^{-4}$---orders of magnitude below the reconstruction loss---so the LUT stays well-behaved without competing with the task, as the operating principle requires.

\subsubsection{SwinLUT Training Loss}

The multi-LUT probe uses the objective introduced in Section~\ref{sec:swinlut}:
\begin{equation}
    \begin{aligned}
    \mathcal{L}_{\text{swin}} = \mathcal{L}_{\text{recon}} &+ \lambda_p \mathcal{L}_{\text{percep}} + \lambda_s \mathcal{L}_{\text{smooth}} \\
    &+ \lambda_m \mathcal{L}_{\text{mono}} + \lambda_e \mathcal{L}_{\text{ent}} + \lambda_d \mathcal{L}_{\text{div}}.
    \end{aligned}
\end{equation}
The smoothness and monotonicity coefficients, $\lambda_s\!=\!10^{-4}$ and $\lambda_m\!=\!10$, are carried over unchanged from the published setting of Zeng~\etal~\cite{zeng2022lut}, whose objective the probe extends; keeping the baseline's values isolates our modifications to the gate.
The large $\lambda_m$ is admissible under the operating principle because monotonicity violations vanish as the bank converges, so $\lambda_m \mathcal{L}_{\text{mono}}$ stays negligible against $\mathcal{L}_{\text{recon}}$---the convergence behavior measured directly for StyleLUTNet above.
The added terms are set as follows.
The perceptual weight $\lambda_p\!=\!0.1$ holds the VGG feature loss~\cite{johnson2016perceptual} an order below reconstruction.
The entropy weight $\lambda_e\!=\!0.1$ matches the auxiliary-loss scale Shazeer~\etal~\cite{shazeer2017moe} report as effective for mixture-of-experts load balancing---the mechanism $\mathcal{L}_{\text{ent}}$ transplants to the LUT gate.
The diversity weight $\lambda_d\!=\!0.01$ is held an order below $\lambda_e$ because diversity contributes little once entropy regularization is active; the ablation of Section~\ref{sec:ablation} probes the contribution of each added term.

\subsubsection{FilmAnalyzerNet Training Loss}

Parameter regression minimizes a weighted sum of squared errors over the 11 outputs of Section~\ref{sec:filmanalyzer}:
\begin{equation}
    \mathcal{L}_{\text{analyzer}} = \sum_{i} w_i\,(\hat{\theta}_i - \theta_i)^2.
    \label{eq:analyzer_loss}
\end{equation}
The output space is heterogeneous---grain intensity lives near $10^{-2}$ while halation radius exceeds 2---so uniform weighting would let large-magnitude parameters dominate the gradient.
The weights compensate: $w\!=\!100$ for grain intensity $\sigma_g$, 50 for halation threshold $t_h$ and intensity $\alpha_h$, 10 for grain size, 5 for each color-bias channel of $\mathbf{b}_h$, and 1 for the luminance-curve coefficients $(a,b,c)$ and halation radius $r$.
They were hand-set to bring the weighted per-parameter errors to comparable magnitude early in training rather than derived from a formal sensitivity analysis; the largest weights fall on the parameters---grain intensity, halation threshold and intensity---whose small absolute errors produce the largest change in the rendered result.

\subsection{Differentiable Trilinear Interpolation}
\label{sec:trilinear}

Both StyleLUTNet and SwinLUT apply their predicted LUTs via the same trilinear interpolation operator.
Each pixel $\mathbf{p} = (r, g, b) \in [0,1]^3$ is treated as a 3D sampling coordinate into the LUT volume.
We implement this through \texttt{F.grid\_sample} with coordinates normalized from $[0,1]$ to $[-1,1]$ and border padding.

This pure-PyTorch operator replaces the custom C++/CUDA extensions of~\cite{zeng2022lut}, which relied on deprecated compiler interfaces incompatible with modern PyTorch ($\geq$2.0) and A100-generation GPUs.
The refactored operator introduces no measurable throughput penalty at the batch sizes used in training ($B\!=\!4$) and supports automatic mixed precision without modification.

\section{Experiments}
\label{sec:experiments}

\subsection{Datasets and Evaluation Protocol}

\paragraph{SwinLUT evaluation.}
MIT-Adobe FiveK~\cite{fivek2011}: 5{,}000 raw photographs retouched by five expert photographers.
We use Expert~C as the target following~\cite{zeng2022lut}, with a split of 3{,}150 training / 350 test images at $480\!\times\!480$ resolution.

\paragraph{StyleLUT + Film Physics evaluation.}
The end-to-end pipeline is evaluated on 350 test image pairs generated through the self-supervised RandomColorTransform procedure (Section~\ref{sec:stylelut}).
Each pair consists of a reference (randomly transformed $I_a$) and target ($I_b$), with the original $I_b$ serving as ground truth.

\paragraph{Protocol rationale.}
No paired dataset exists in which arbitrary photographs are graded to match specific film stocks; constructing one would require professional color grading of every input against every stock.
The self-supervised protocol substitutes a measurable surrogate: it admits an exact ground truth (the untransformed $I_b$), so PSNR and SSIM are computed against a known target rather than a subjective ideal.
The substitution defines what the numbers certify---the model's ability to invert color transforms---and what they do not, namely fidelity to a real emulsion's spectral response, which the expert evaluation (Section~\ref{sec:user_study}) assesses instead.

\paragraph{Metrics.}
We report peak signal-to-noise ratio (PSNR), structural similarity (SSIM)~\cite{wang2004ssim}, and learned perceptual image patch similarity (LPIPS)~\cite{zhang2018lpips}.
PSNR measures pixel fidelity; SSIM measures structural agreement under luminance and contrast normalization; LPIPS measures distance in a learned feature space aligned with human perception.
The three diverge on the optical stages by construction: grain and halation are deliberate distortions, so they lower PSNR and raise LPIPS even when they improve realism (Section~\ref{sec:metric_limits}).
We therefore read pixel metrics as evidence for the color stage and defer perceptual judgment of the full pipeline to the expert evaluation (Section~\ref{sec:user_study}).

\subsection{Implementation Details}

\paragraph{SwinLUT.}
28.3M parameters (predominantly Swin-T backbone).
AdamW with group-specific learning rates: $5\!\times\!10^{-5}$ (backbone), $5\!\times\!10^{-4}$ (weight head), $10^{-3}$ (LUT parameters and $\tau$).
Weight decay $5\!\times\!10^{-4}$; cosine annealing with warm restarts ($T_0\!=\!50$, $T_{\text{mult}}\!=\!1$, $\eta_{\min}\!=\!10^{-6}$).
Mixed-precision (FP16) training, batch size 4.
Data augmentation: paired random crops ($552 \to 480$), random horizontal flip, input-only brightness jitter.
Early stopping with patience 25.

\paragraph{StyleLUTNet.}
ResNet-18 encoder (11.2M parameters) + MLP decoder (71.4M parameters), for 82.6M parameters in total.
AdamW, learning rate $10^{-4}$ for encoder (fine-tuned), $5\!\times\!10^{-4}$ for decoder.
Trained for 100 epochs on RandomColorTransform pairs; best model selected by validation loss (validation PSNR 20.19~dB under the strong color distortions of the training pairs---a monitoring figure; test-set results on separately generated pairs appear in Section~\ref{sec:e2e_eval}).
Training time: 1 hour 42 minutes on a single NVIDIA A100 (RIT SPORC cluster).

\paragraph{FilmAnalyzerNet.}
ResNet-18 backbone with seven regression heads.
AdamW, learning rate $10^{-4}$, trained for 50 epochs.
5\% of training pairs are identity transforms (reference equals target) to anchor the model's zero-distortion prediction.

All training performed on the RIT SPORC computing cluster using NVIDIA A100 GPUs with 80~GB memory.

\subsection{Diagnosing the Multi-LUT Collapse}
\label{sec:swinlut_dev}

Three configurations of the multi-LUT model isolate the collapse, its mitigation, and the generalization behavior that follows.

\subsubsection{V1: Baseline Swin-T + 3D LUT}

The initial model replaced ResNet-18 with Swin-T while retaining the original recipe: standard softmax (no temperature), three loss terms (L1, smoothness, monotonicity), identity initialization for only the first basis LUT.

\paragraph{LUT weight collapse.}
Training converged to 21.27~dB PSNR and 0.9026 SSIM.
The softmax output collapsed onto a single basis LUT: mean per-LUT weights over the 350 test images are $[0.947,\,0.050,\,0.002,\,0.001,\,0.001]$, and mean weight entropy is 0.16 of a possible 1.61---under 10\% of the gate's capacity in use.
Four of five basis LUTs were effectively ignored.

\begin{figure*}[t]
  \centering
  \includegraphics[width=\textwidth]{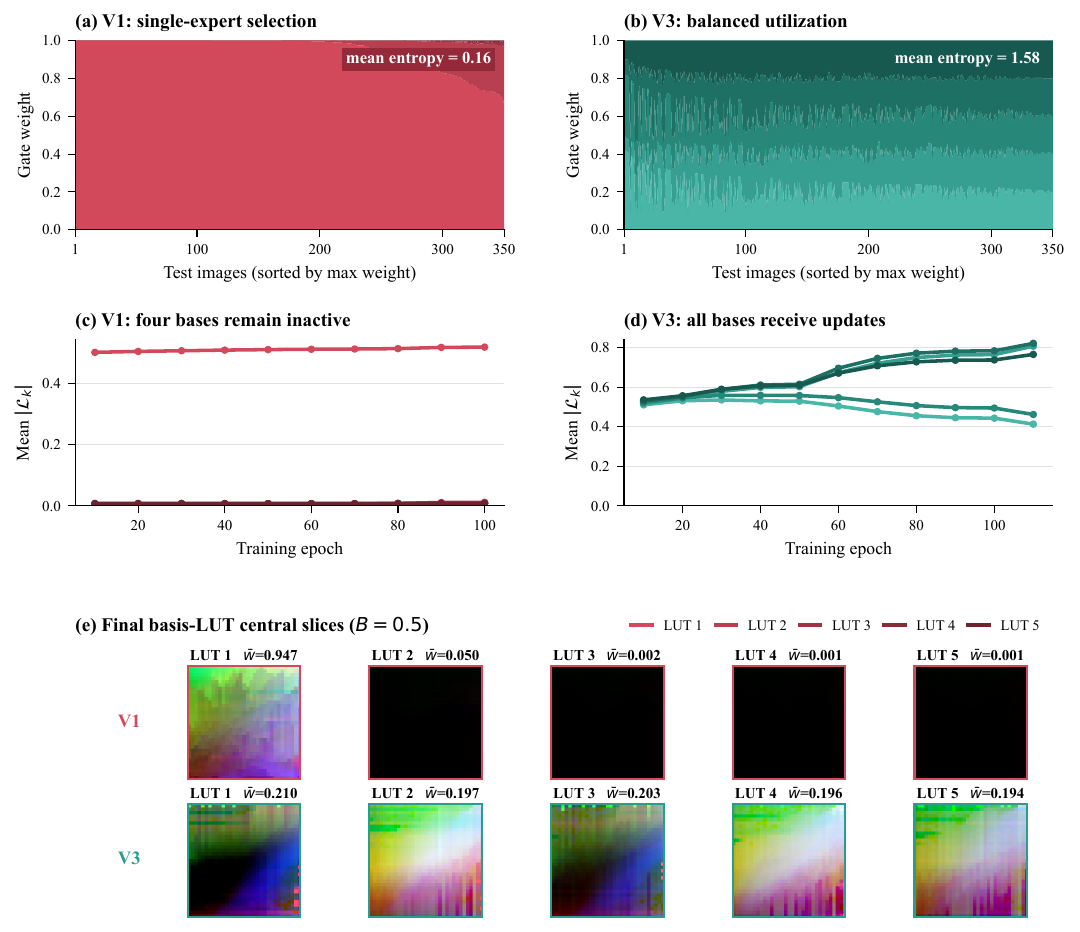}
  \caption{Anatomy of multi-LUT weight collapse. (a--b)~Gate distributions over all 350 FiveK test images. V1 assigns a mean weight of 0.947 to LUT~1 (entropy 0.16), whereas entropy-regularized V3 uses all five bases (entropy 1.58; maximum 1.61). (c--d)~Checkpoint-wise mean absolute LUT coefficients at 10-epoch intervals. In V1, only LUT~1 develops substantial magnitude while four bases remain effectively inactive; in V3, every basis evolves during training. (e)~Central $B=0.5$ slices of the final basis LUTs, with mean test-set gate weights $\bar{w}$. The suppressed V1 bases remain nearly black, while V3 learns five distinct color mappings.}
  \label{fig:weight-distribution}
\end{figure*}

This collapse arises from the interaction between softmax dynamics and end-to-end gradient flow.
Early in training, slight differences in gradient magnitude across LUTs cause the optimizer to preferentially improve one.
As that LUT becomes more accurate, the weight predictor assigns it higher probability, concentrating more gradient signal on it---a positive feedback loop.
Without explicit pressure toward diversity, the system converges to a degenerate single-LUT solution.
Figure~\ref{fig:weight-distribution} exposes both consequences of this feedback loop: the gate selects one expert across the test set, and the four suppressed basis LUTs remain near their zero initialization throughout training.

\subsubsection{V2: Mitigating Weight Collapse}

V2 introduced: (1)~temperature-scaled softmax with learnable $\tau$; (2)~entropy regularization on predicted weights; (3)~pairwise diversity loss between basis LUTs; (4)~symmetric identity initialization of all five LUTs with scaled perturbation; (5)~a VGG-based perceptual loss~\cite{simonyan2015vgg,johnson2016perceptual}.
The weight head was deepened to three layers with LayerNorm and GELU.

V2 reached 22.19~dB PSNR and 0.9289 SSIM---gains of +0.92~dB and +0.026 over V1.
Weight entropy rose to 1.588 (of maximum $\ln 5 = 1.609$).
The dominant weight dropped from 0.95 to 0.208.

The train--test gap, however, reached 3.77~dB (train 25.68 vs.\ test 21.92 at the final epoch), signaling overfitting on a training set of 3{,}150 images.

\subsubsection{V3: Regularization}

V3 targeted overfitting: (1)~paired data augmentation; (2)~backbone drop path rate raised to 0.2; (3)~weight decay increased to $5\!\times\!10^{-4}$; (4)~early stopping with patience 25.

Training early-stopped at epoch 114 (best at epoch 89).
Test PSNR: 22.16~dB; SSIM: 0.9270; weight entropy: 1.577.
The generalization gap decreased to 2.27~dB---a 40\% reduction from V2.

\subsection{SwinLUT Version Comparison}

\begin{table}[htbp]
\centering
\small
\caption{Progression across SwinLUT model versions on MIT-Adobe FiveK (Expert~C, 350 test images).}
\label{tab:comparison}
\begin{tabular}{@{}lccc@{}}
\toprule
\textbf{Metric} & \textbf{V1} & \textbf{V2} & \textbf{V3} \\
\midrule
Test PSNR (dB)         & 21.27  & 22.19  & 22.16 \\
Test SSIM              & 0.903  & 0.929  & 0.927 \\
Weight Entropy         & 0.158  & 1.588  & 1.577 \\
Overfit Gap (dB)       & 0.57   & 3.77   & 2.27 \\
Dominant LUT Wt.       & 0.947  & 0.208  & 0.263 \\
\midrule
Data Augmentation      & ---    & ---    & \checkmark \\
Perceptual Loss        & ---    & \checkmark & \checkmark \\
Entropy Reg.           & ---    & \checkmark & \checkmark \\
Diversity Loss         & ---    & \checkmark & \checkmark \\
Temp-scaled softmax    & ---    & \checkmark & \checkmark \\
Early Stopping         & ---    & ---    & \checkmark \\
Drop Path Rate         & 0.0   & 0.0    & 0.2 \\
Weight Decay           & $10^{-4}$ & $10^{-4}$ & $5\!\times\!10^{-4}$ \\
\bottomrule
\end{tabular}
\end{table}

\subsection{SwinLUT Ablation Study}
\label{sec:ablation}

Four ablation variants were trained for 30 epochs each using the V3 architecture, disabling one or more loss terms per variant.
Results appear in Table~\ref{tab:ablation}; the fully trained V3 model is included as a reference row, but its longer schedule means comparisons against it conflate a term's effect with training budget, so the analysis below relies on equal-budget contrasts among the 30-epoch variants.

\begin{table}[htbp]
\centering
\small
\caption{Ablation study: effect of disabling loss components in SwinLUT. Ablation variants trained 30 epochs with the V3 architecture; best validation metrics reported. Full V3 is the fully trained model of Table~\ref{tab:comparison}, shown for reference.}
\label{tab:ablation}
\begin{tabular}{@{}lccc@{}}
\toprule
\textbf{Configuration} & \textbf{PSNR} & \textbf{SSIM} & \textbf{Entropy} \\
\midrule
Full V3                & \textbf{22.16}  & \textbf{0.927} & 1.577 \\
No Perceptual          & 21.86  & 0.922 & 1.569 \\
No Entropy Reg.        & 20.96  & 0.905 & 0.000 \\
No Diversity           & 21.70  & 0.917 & 1.571 \\
L1 Only                & 20.89  & 0.905 & 0.000 \\
\bottomrule
\end{tabular}
\end{table}

\paragraph{Entropy regularization is the dominant factor.}
Removing it causes complete weight collapse: entropy falls to zero, the dominant weight reaches 1.0, and PSNR drops to 20.96~dB---nearly identical to the L1-only baseline.
Weight collapse is the default outcome of unconstrained softmax training in multi-LUT architectures; entropy regularization is necessary and, with temperature scaling, sufficient to prevent it in our experiments.

\paragraph{Diversity loss is marginal when entropy regularization is active.}
Removing diversity while retaining entropy leaves weight entropy essentially unchanged (1.571): full utilization does not depend on the explicit diversity term.
The equal-budget contrast makes the point directly---the two variants that differ by which gate term they keep, No~Entropy (diversity only, 20.96~dB) and No~Diversity (entropy only, 21.70~dB), differ by 0.74~dB in entropy's favor.
If the optimizer must use all LUTs (enforced by entropy regularization), it has intrinsic incentive to differentiate them.

\paragraph{The perceptual term's contribution is not separable at this budget.}
The strongest 30-epoch variant is the one without the perceptual loss (21.86~dB), and adding the term to the collapsed configurations changes only 0.07~dB (No~Entropy 20.96 vs.\ L1~Only 20.89).
The 0.29~dB margin of the fully trained model over the No-Perceptual variant conflates the term's effect with the longer schedule; the ablation therefore does not isolate a clear PSNR benefit, and the term is retained from the V2 recipe (Section~\ref{sec:swinlut_dev}).

\subsection{StyleLUT Regularization Collapse}
\label{sec:stylelut_training}

The StyleLUTNet training revealed a distinct failure mode from the weight collapse observed in SwinLUT.
When smoothness and monotonicity loss weights were set to $\lambda_s\!=\!10$ and $\lambda_m\!=\!5$ (a combined regularization-to-pixel ratio of 15:1), the learnable residual scale parameter $\alpha$ shrank from its initial value of 0.1 to 0.04 over 100 epochs.
The network produced near-identity output: the predicted LUT deviated minimally from the identity mapping because the optimizer discovered that reducing $\alpha$ was the most efficient path to lowering the smoothness penalty.

The fix was twofold: (1)~convert $\alpha$ from \texttt{nn.Parameter} to a fixed buffer at 1.0, and (2)~reduce regularization weights to $\lambda_s\!=\!1.0$, $\lambda_m\!=\!0.5$ while increasing histogram loss to $\lambda_h\!=\!3.0$.
After retraining, smoothness and monotonicity losses converged to approximately $10^{-4}$---confirming that the LUT remained well-conditioned without aggressive regularization---and color transfer quality improved from near-identity to visually accurate reproduction of the target style.

This failure mode is distinct from LUT weight collapse (Section~\ref{sec:swinlut}) but shares a root cause: when auxiliary losses dominate reconstruction, the optimizer finds degenerate solutions that minimize auxiliary terms at the expense of the primary task.
The practical lesson is that LUT regularization terms should be several orders of magnitude below the reconstruction loss in the converged regime.

\subsection{End-to-End Pipeline Evaluation}
\label{sec:e2e_eval}

Table~\ref{tab:pipeline} reports metrics as each pipeline stage is added.
The progression reveals the expected pattern: pixel-level metrics (PSNR, SSIM) decrease slightly as grain and halation introduce intentional distortion, while perceptual distance (LPIPS) increases because the added texture diverges from the clean ground truth.

\begin{table}[htbp]
\centering
\small
\caption{End-to-end pipeline ablation. Each row adds one processing stage to the previous. Evaluated on 350 self-supervised test pairs. V6 jointly fine-tunes the stock-conditioned SwinLUT, GrainNet, and HalationNet predecessor modules; earlier rows use independently trained modules.}
\label{tab:pipeline}
\begin{tabular}{@{}lccc@{}}
\toprule
\textbf{Configuration} & \textbf{PSNR$\uparrow$} & \textbf{SSIM$\uparrow$} & \textbf{LPIPS$\downarrow$} \\
\midrule
Naive (no processing)   & 21.19  & 0.915  & 0.517 \\
StyleLUT only            & \textbf{22.05}  & \textbf{0.925}  & \textbf{0.308} \\
+ Grain                  & 21.82  & 0.923  & 0.415 \\
+ Grain + Halation       & 21.76  & 0.923  & 0.416 \\
V6 (E2E, all stages)     & 21.72  & 0.923  & 0.416 \\
\bottomrule
\end{tabular}
\end{table}

\paragraph{StyleLUT accounts for the majority of improvement.}
Color grading alone raises PSNR by +0.86~dB and reduces LPIPS by 40\% relative to the naive baseline, confirming that the conditional LUT captures the dominant component of the target style.

\paragraph{Grain reduces PSNR but adds perceptual realism.}
Adding grain drops PSNR by 0.23~dB, which is expected---grain is intentional noise.
SSIM remains nearly unchanged (0.925 $\to$ 0.923), indicating that structural content is preserved.
The LPIPS increase from 0.308 to 0.415 reflects the perceptual model's sensitivity to high-frequency texture changes, but does not indicate quality degradation: the added grain is a desired visual effect.

\paragraph{Halation is subtle but measurable.}
The halation stage contributes a further 0.06~dB PSNR decrease (21.82 $\to$ 21.76), consistent with its physically localized effect on highlight regions.

\paragraph{Joint training preserves reconstruction quality.}
The stock-conditioned V6 predecessor jointly fine-tunes its color, grain, and halation modules and achieves 21.72~dB---only 0.04~dB below the separately trained pipeline.
This result shows that differentiable optical rendering can be optimized jointly with LUT-based color processing without materially reducing reconstruction quality; it is not a direct parameter-recovery evaluation of the open-set FilmAnalyzerNet described in Section~\ref{sec:filmanalyzer}.

\subsection{Inference Efficiency}
\label{sec:efficiency}

\begin{figure}[tbp]
  \centering
  \includegraphics[width=\columnwidth]{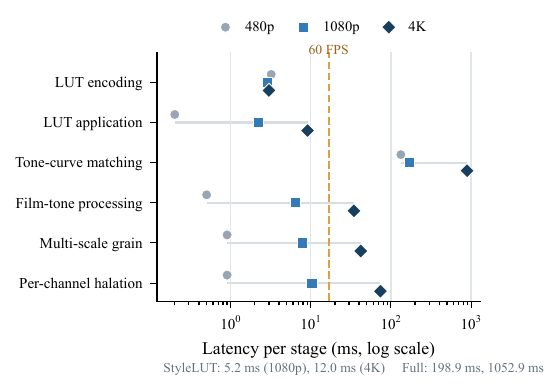}
  \caption{Per-stage inference latency on an NVIDIA RTX~4060 GPU (median of 20 runs). The logarithmic axis separates the resolution-independent LUT encoder from pixel-count-dependent stages. The dashed line marks the 16.7~ms budget for 60~FPS. Tone-curve matching is the dominant bottleneck, while StyleLUT encoding and application total 5.2~ms at 1080p and 12.0~ms at 4K.}
  \label{fig:speed}
\end{figure}

\begin{table}[htbp]
\centering
\caption{Cross-device comparison at 1080p and 4K (ms). RTX~4060 provides the fastest absolute times; Apple~Silicon MPS on a MacBook~Pro (2023) offers a portable alternative.}
\label{tab:speed_device}
\resizebox{\columnwidth}{!}{
\begin{tabular}{@{}lcccc@{}}
\toprule
 & \multicolumn{2}{c}{\textbf{1080p}} & \multicolumn{2}{c}{\textbf{4K}} \\
\cmidrule(lr){2-3}\cmidrule(lr){4-5}
\textbf{Device} & StyleLUT & Full & StyleLUT & Full \\
\midrule
RTX 4060 (CUDA) & 5.2 & 198.9 & 12.0 & 1052.9 \\
Apple MPS       & 11.1 & 510.1 & 27.7 & 1803.4 \\
CPU (Apple M-series) & 102.0 & 1783.0 & 359.4 & 6323.7 \\
\bottomrule
\end{tabular}
}
\end{table}

Figure~\ref{fig:speed} breaks down per-stage latency on an RTX~4060. The StyleLUT color grading path---LUT encoding via ResNet-18 followed by trilinear interpolation---runs in \textbf{5.2\,ms at 1080p} (192\,FPS) and 12.0\,ms at 4K (83\,FPS), confirming that the core neural-network component is suitable for real-time, interactive use. LUT encoding is resolution-independent at ${\sim}3$\,ms since the encoder operates on a fixed $224\!\times\!224$ reference crop; only the trilinear application step scales with output resolution.

The full pipeline is dominated by CDF-based tone curve matching, which accounts for 85\% of total latency at 1080p. This bottleneck is algorithmic rather than architectural: the current implementation iterates over 256 histogram bins per channel with per-pixel tensor comparisons. Replacing this loop with \texttt{torch.histc} or a compiled CUDA kernel would substantially reduce this cost; we leave this optimization to future work.

Table~\ref{tab:speed_device} compares three hardware configurations. The RTX~4060 is $2.1\times$ faster than Apple~MPS for StyleLUT and $2.6\times$ faster for the full pipeline at 1080p. Apple~MPS still achieves 11.1\,ms StyleLUT latency---adequate for interactive previewing on a laptop. CPU-only inference is $20\times$ slower for StyleLUT, confirming the importance of GPU acceleration even on Apple~Silicon.

A key practical advantage is that the predicted 3D LUT can be exported as a standard \texttt{.cube} file and applied natively in Adobe Photoshop, DaVinci Resolve, or Capture One at zero neural-network cost. For photographers who want the color grading without the film physics emulation, the exported LUT provides the result at the speed of the host application's built-in LUT engine---effectively free.

\subsection{Qualitative Results}
\label{sec:qualitative}

\begin{figure*}[t]
  \centering
  \includegraphics[width=\textwidth]{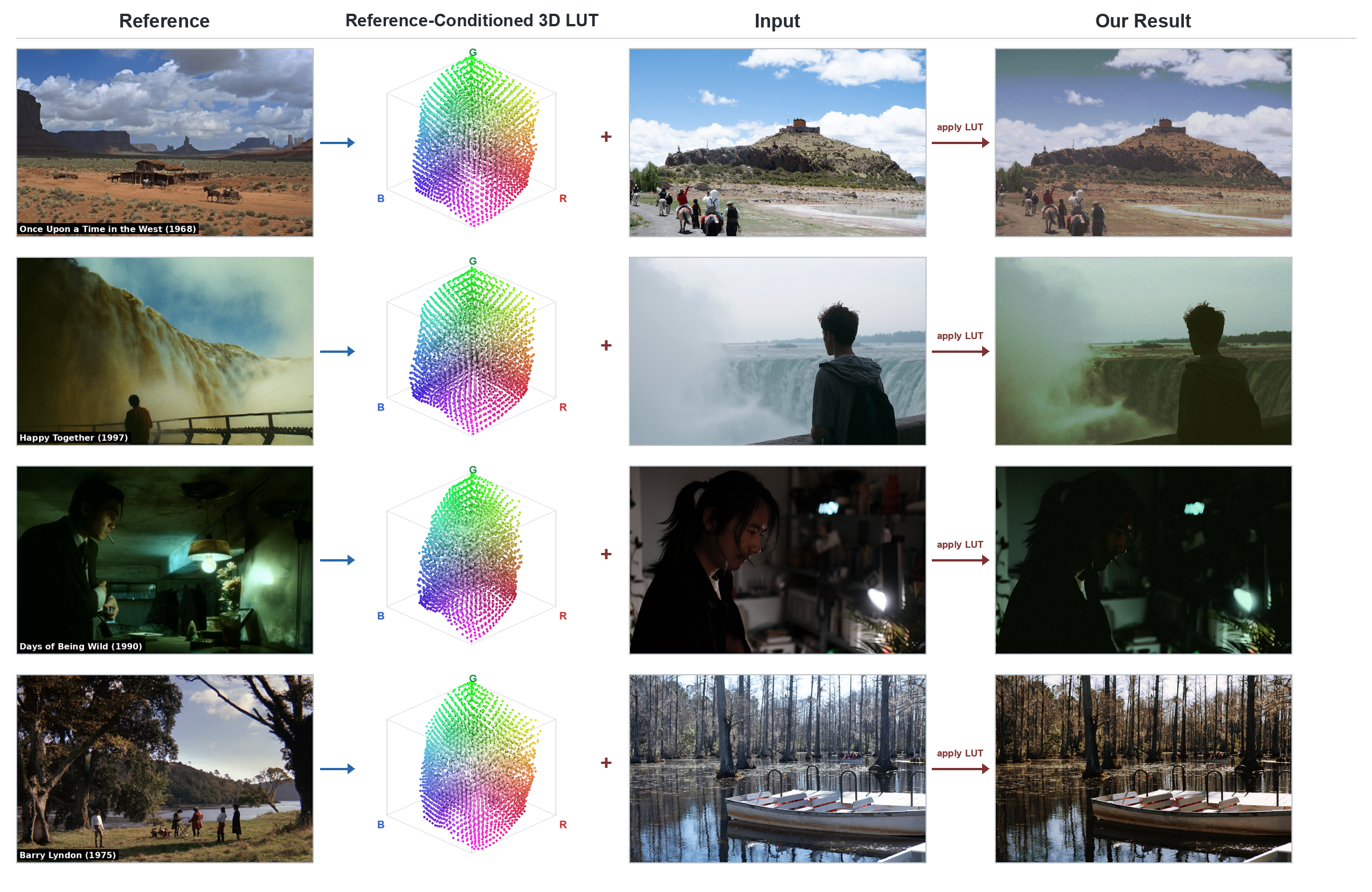}
  \caption{Open-set generalization across diverse film stocks and scene types. Each row shows a reference film scan, its reference-conditioned 3D LUT, a digital input photograph, and the Deep Analog output after applying the inferred look. Point positions and colors in each LUT visualization encode output RGB coordinates. References, top to bottom: \textit{Once Upon a Time in the West} (1968), \textit{Happy Together} (1997), \textit{Days of Being Wild} (1990), \textit{Barry Lyndon} (1975). Digital inputs are original photographs by the author. The system reproduces each reference's characteristic color palette and tonal character without film-stock-specific training.}
  \label{fig:qualitative}
\end{figure*}

\begin{figure*}[t]
  \centering
  \includegraphics[width=\textwidth]{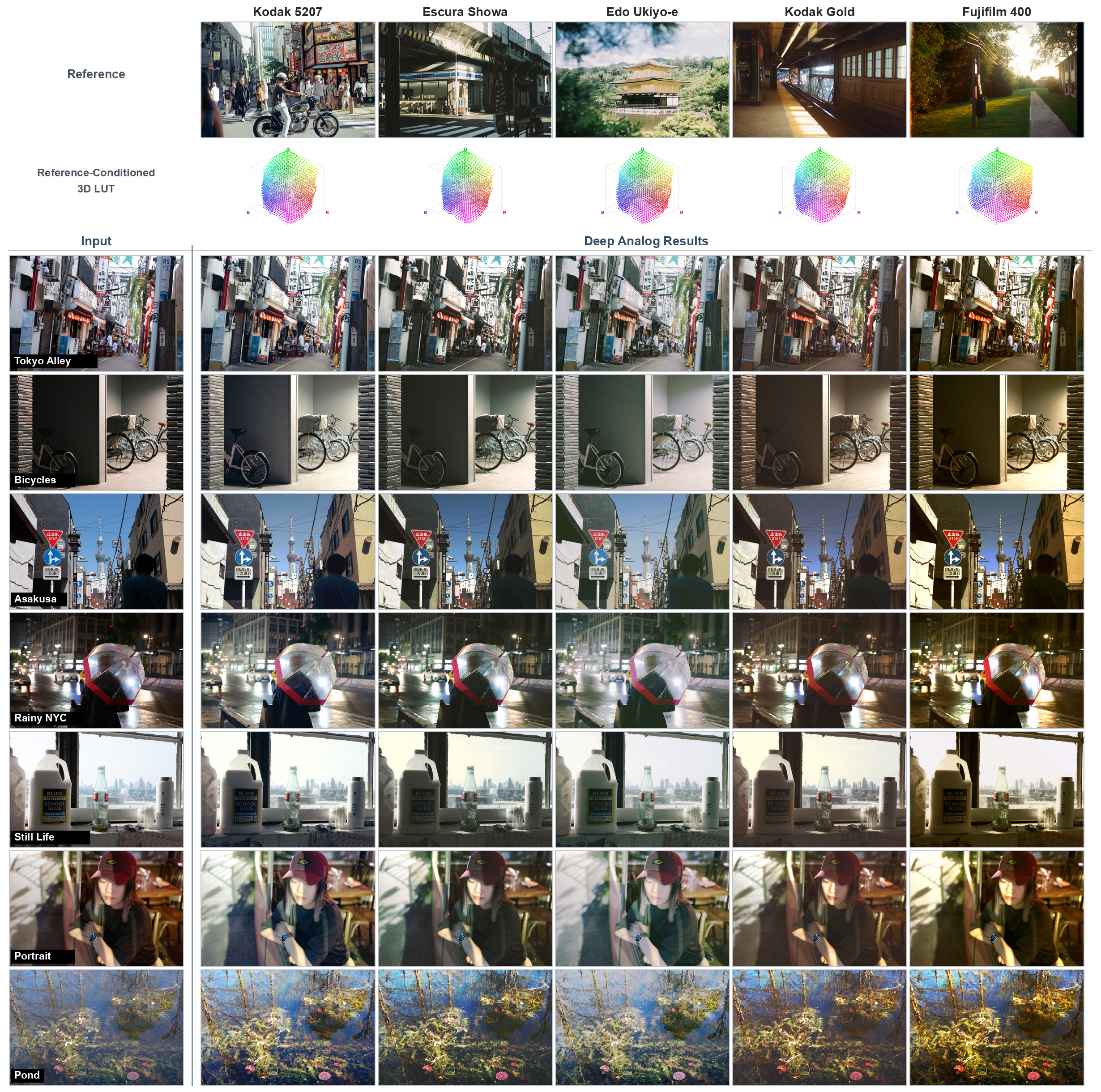}
  \caption{Multi-reference film look transfer across five film stocks and seven input scenes. The top two rows show each uncropped reference film scan and its corresponding reference-conditioned 3D LUT; point positions and colors encode LUT output RGB coordinates. References were captured by the author on Kodak Vision3 5207 250D, Escura Showa Camera, Escura Edo Ukiyo-e 400, Kodak Gold 400, and Fujifilm 400. The separated leftmost column contains uncropped digital inputs spanning daylight, low-light, indoor, portrait, and nature conditions. Each result cell preserves the source aspect ratio and shows the Deep Analog output conditioned on the corresponding reference---all produced by a single model without film-stock-specific fine-tuning. All photographs are by the author.}
  \label{fig:film_matrix}
\end{figure*}

\begin{figure*}[t]
  \centering
  \includegraphics[width=\textwidth]{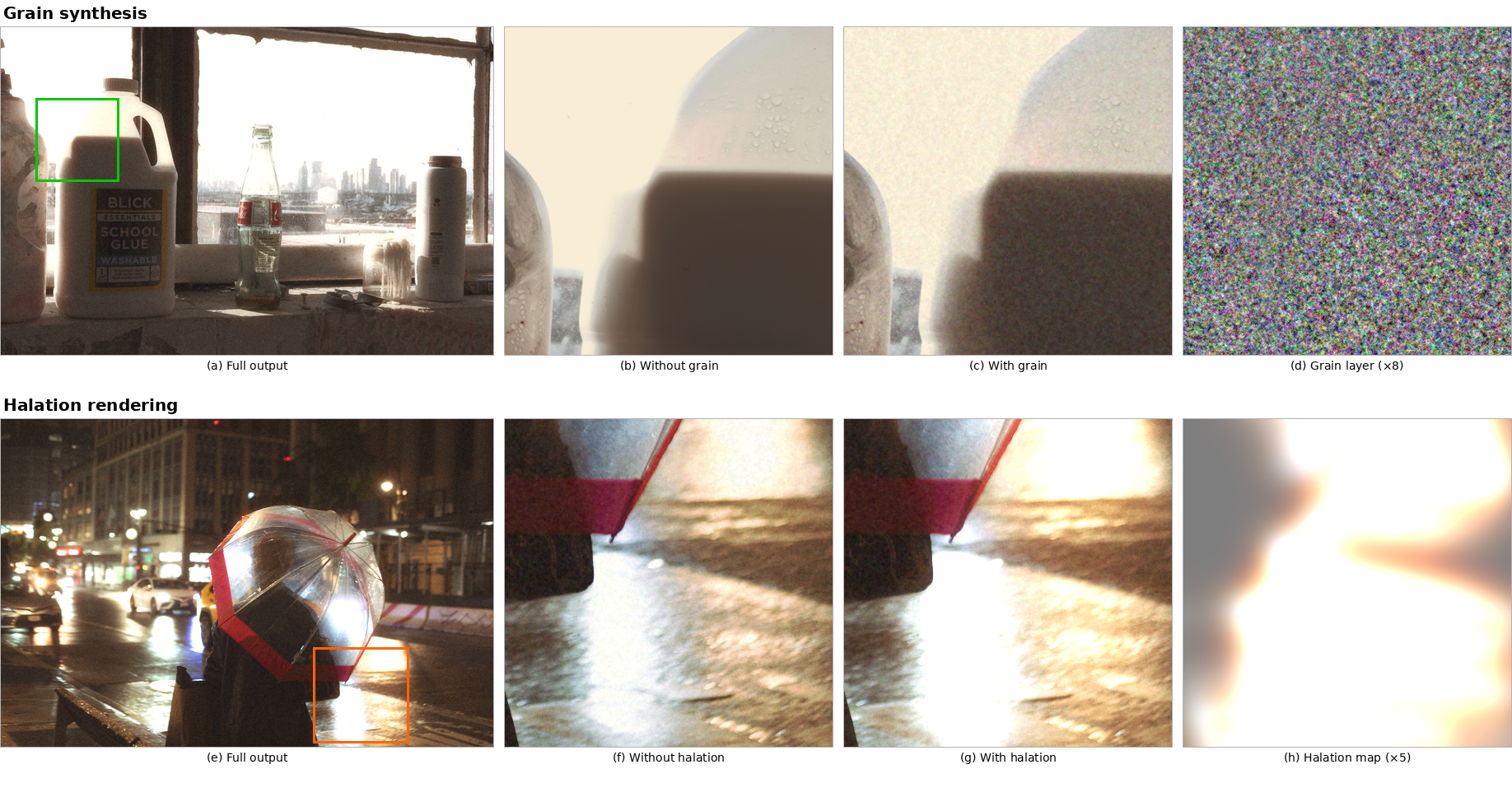}
  \caption{Close-up comparison of grain synthesis and halation rendering. \textbf{Top row (grain):} (a)~Full output with crop region marked in green, (b)~a white bottle surface before grain, (c)~the same region after grain synthesis---note the fine noise texture on the previously smooth surface, (d)~grain layer amplified $8\times$, showing luminance-dependent density variation. \textbf{Bottom row (halation):} (e)~Full output (rainy night scene) with crop region marked in orange, (f)~wet pavement reflections before halation, (g)~after halation---the bright reflections develop a warm glow that spreads across the road surface, (h)~halation map amplified $5\times$, revealing the per-channel color spread: the red channel, which dominates the base-reflected re-exposure, spreads with a larger radius than blue. Halation parameters are amplified ${\sim}3\times$ relative to defaults for print visibility. Both images: digital photographs by the author; reference: Kodak Gold 400 film scan.}
  \label{fig:grain_halation}
\end{figure*}

\paragraph{Color transfer fidelity.}
Figure~\ref{fig:qualitative} shows the core capability of StyleLUTNet: given a single reference film scan, the system produces a color-graded output that captures the reference's characteristic palette and tonal distribution. The warm amber tones of \textit{Once Upon a Time in the West}, the teal-and-gold cross-processing of \textit{Happy Together}, the desaturated greens of \textit{Days of Being Wild}, and the candlelight warmth of \textit{Barry Lyndon} are each reproduced from a single reference image without any film-stock-specific training. Because the model predicts a full 3D LUT conditioned on the reference, it applies a globally coherent color transformation rather than a per-pixel mapping, which preserves spatial structure and avoids local artifacts.

\paragraph{Generalization across scenes and stocks.}
Figure~\ref{fig:film_matrix} tests generalization along two axes simultaneously: five distinct film stocks (Kodak Vision3 5207 250D, Escura Showa, Escura Edo Ukiyo-e 400, Kodak Gold 400, Fujifilm 400) and seven input scenes spanning daylight, low-light, indoor, portrait, and nature conditions. Every cell in the $5\!\times\!7$ grid is produced by the same model with the same weights---only the reference changes. Two patterns are worth noting. First, the color identity of each film stock is consistent across scenes: Kodak Gold 400 produces its signature warm yellows regardless of whether the input is an outdoor street or an indoor close-up. Second, the model handles challenging illumination gracefully: low-light and mixed-lighting inputs receive plausible color grading rather than saturated noise or washed-out tones.

\paragraph{Film physics effects.}
Figure~\ref{fig:grain_halation} isolates the two film physics stages---grain synthesis and halation rendering---by comparing crops extracted before and after each stage. In the grain comparison (top row), a white bottle surface that appears smooth without grain acquires fine multi-scale noise after the grain stage; the amplified grain layer~(d) shows that grain density varies with luminance, matching the physical behavior of silver halide crystals: denser in mid-tones, sparser in deep shadows and saturated highlights. In the halation comparison (bottom row), bright reflections on wet pavement acquire a warm glow that spreads well beyond the original highlight boundaries---a characteristic signature of real film, where light penetrates the emulsion, reflects off the base, and re-exposes it from behind. The amplified halation map~(h) makes the per-channel spread visible: because the red-sensitive layer lies deepest and dominates the base-reflected re-exposure, the halo spreads widest in red, producing the warm tone that distinguishes halation from simple Gaussian blur.

\subsection{Comparison with Baselines}
\label{sec:baseline_comparison}

Table~\ref{tab:baseline} compares Deep Analog against three families of LUT-based image enhancement methods.
The comparison reveals a structural divide: Zeng~\etal~\cite{zeng2022lut}, AdaInt~\cite{yang2022adaint}, and SepLUT~\cite{yang2022seplut} are \emph{closed-set} methods that learn a single fixed mapping from input images to a predetermined target style (Expert~C retouching in FiveK).
StarEnhancer~\cite{song2021star} extends this paradigm by conditioning on a learned style embedding, enabling multi-style enhancement from a single model, but the set of styles remains fixed at training time.
StyleLUTNet is fully \emph{open-set}: it accepts an arbitrary reference photograph and predicts a conditional 3D LUT to match that reference's color characteristics, requiring no retraining or fine-tuning.

\begin{table}[htbp]
\centering
\caption{Comparison with LUT-based baselines. Closed-set methods learn a fixed enhancement mapping (FiveK Expert~C, 480p). StyleLUTNet is open-set, evaluated on self-supervised pairs ($\dagger$). Baseline values are from the respective publications.}
\label{tab:baseline}
\resizebox{\columnwidth}{!}{
\begin{tabular}{@{}lcclcc@{}}
\toprule
\textbf{Method} & \textbf{Params} & \textbf{LUTs} & \textbf{Conditioning} & \textbf{PSNR} & \textbf{SSIM} \\
\midrule
3D LUT~\cite{zeng2022lut}  & 594K & 3$\times$33$^3$ & None (fixed) & 25.28 & 0.926 \\
AdaInt~\cite{yang2022adaint} & 620K & 3$\times$33$^3$ & None (fixed) & 25.50 & 0.930 \\
SepLUT~\cite{yang2022seplut} & $\sim$280K & 3$\times$1D & None (fixed) & --- & --- \\
StarEnhancer~\cite{song2021star} & --- & 3$\times$33$^3$ & Style embedding & --- & --- \\
\midrule
SwinLUT V3 (ours) & 28.3M & 5$\times$33$^3$ & None (fixed) & 22.16 & 0.927 \\
StyleLUTNet (ours) & 82.6M & 1$\times$33$^3$ & Ref.\ image & 22.05$^\dagger$ & 0.925$^\dagger$ \\
\bottomrule
\end{tabular}
}
\end{table}

\paragraph{PSNR gap reflects different objectives, not architectural weakness.}
On FiveK Expert~C, the closed-set baselines achieve 25--26~dB PSNR with lightweight CNN backbones under 1M parameters.
Our SwinLUT~V3, trained on the same benchmark, reaches 22.16~dB---a gap of roughly 3~dB.
Two factors account for most of this difference.
First, SwinLUT was an exploratory architecture designed to study weight collapse in multi-LUT systems (Section~\ref{sec:ablation}), not to maximize FiveK accuracy; it was trained with aggressive regularization (entropy, diversity, and perceptual losses) and early-stopped at epoch 114 (Section~\ref{sec:swinlut_dev}), whereas the baselines are fully optimized for reconstruction fidelity on FiveK over longer schedules.
Second, the Swin-T backbone introduces 28.3M parameters---nearly 50$\times$ the baselines' sub-million-parameter CNNs---yet produces only five scalar weights, creating a severe bottleneck between encoder capacity and LUT output dimensionality.
The progression from SwinLUT to StyleLUTNet eliminated this bottleneck by replacing the multi-LUT blending with a single conditional LUT predicted end-to-end (Section~\ref{sec:stylelut}).

StyleLUTNet's 22.05~dB is measured on a different task entirely: self-supervised color transfer where ground truth is generated by applying random color distortions.
Comparing this number against FiveK baselines is not meaningful, because StyleLUTNet never sees FiveK training pairs and is not optimizing for a fixed expert's retouching style.

\paragraph{Inference speed.}
All LUT-based methods share a common computational profile: a lightweight encoder predicts LUT parameters, and the LUT itself is applied via trilinear interpolation at negligible cost.
Zeng~\etal~\cite{zeng2022lut} report under 2~ms to process a 4K image on a Titan~RTX; StarEnhancer~\cite{song2021star} reports over 200~FPS at 4K resolution.
Our StyleLUT color grading path runs in 5.2\,ms at 1080p and 12.0\,ms at 4K on an RTX~4060 (Figure~\ref{fig:speed}), which is slower than the baselines primarily because the ResNet-18 encoder operates on $224\!\times\!224$ reference images rather than the $256\!\times\!256$ thumbnails used by the baseline CNNs, and because the decoder MLP alone contains 71.4M parameters (82.6M for the full model).
At 192~FPS (1080p), StyleLUT remains well within the real-time threshold despite its larger model size.

\paragraph{The open-set advantage.}
The critical distinction is not PSNR on FiveK but what each method can do in practice.
The baselines require paired training data for each target style: to emulate Kodak Vision3 250D, one would need thousands of input--target image pairs graded to match that specific film stock.
No such paired dataset exists for the film emulation task, and constructing one would require professional color grading of each input to match each film stock---an impractical undertaking for more than a handful of styles.
StyleLUTNet sidesteps this requirement entirely: the user supplies a single reference frame, and the model predicts the color transformation at test time.
This open-set capability is the primary contribution of the architecture, and it comes with a practical deployment advantage: the predicted LUT can be exported as a standard \texttt{.cube} file for zero-cost application in any professional editing tool (Section~\ref{sec:efficiency}).

\subsection{Expert Evaluation}
\label{sec:user_study}

Because no reference-based metric scores the full pipeline correctly (Section~\ref{sec:metric_limits}), we ran a small expert evaluation to supply the perceptual judgment the metrics cannot.
We recruited four evaluators with professional backgrounds in color and imaging science, each self-reporting expertise in color science and photography (three also in imaging and image processing, two in film and analog media).
The panel is small, so we report the study as a pilot: it indicates direction and surfaces the failure modes experts notice, but the sample supports no claim of statistical significance, and we report no confidence intervals.

The instrument has two parts.
Part~1 (\emph{reference match}) presents Deep Analog outputs beside the reference film scan and asks how well the output matches the reference stock on a 1--5 scale, over a $7\!\times\!5$ grid of scenes and stocks; five cells were dropped for missing stimulus files, leaving 30 items and 120 ratings.
Part~2 (\emph{real-or-rendered}) is a forced-choice discrimination: each of 16 trials pairs a real film photograph with a Deep Analog rendering, and the evaluator selects which is the real film---a Turing-style test where an ideal emulator drives accuracy toward the 50\% chance rate.
Table~\ref{tab:userstudy} summarizes both parts.

\begin{table}[htbp]
\centering
\small
\caption{Expert evaluation ($n\!=\!4$ imaging professionals). Part~1: mean reference-match rating (1--5). Part~2: accuracy at telling real film from Deep Analog output; 50\% is chance. With only four evaluators, counts are reported descriptively, without an inferential test.}
\label{tab:userstudy}
\begin{tabular}{@{}lcc@{}}
\toprule
\textbf{Evaluator} & \textbf{Part 1 mean} & \textbf{Part 2 acc.} \\
\midrule
E1 & 3.17 & 10/16 \\
E2 & 3.03 & \phantom{0}9/16 \\
E3 & 3.60 & 10/16 \\
E4 & 3.23 & \phantom{0}9/16 \\
\midrule
Pooled & 3.26 (SD 0.94) & 38/64 = 59.4\% \\
\bottomrule
\end{tabular}
\end{table}

\paragraph{Reference match.}
Pooled Part~1 ratings average 3.26 of 5 (SD~0.94), with 81\% of ratings at or above the ``moderate match'' midpoint and the four evaluators closely aligned (means 3.17, 3.03, 3.60, 3.23).
Per-stock means separate the results along the axis the design targets: the desaturated Escura Showa look scores highest (3.75) and the saturated, high-key Edo Ukiyo-e lowest (2.96), consistent with the boundary cases of Section~\ref{sec:boundary}, where strong-color references over-drive low-saturation scenes.

\paragraph{Real-or-rendered discrimination.}
Pooled Part~2 accuracy is 59.4\% (38/64), individual scores 10, 9, 10, and 9 of 16---all close to the 50\% chance rate.
Descriptively, even evaluators trained in color and imaging science were near chance at telling Deep Analog output from real film.
This is the study's strongest signal, and it is the right target---film emulation succeeds when the rendering is \emph{indistinguishable} from film, not when it is preferred over a baseline.

\paragraph{What to improve.}
Asked for the weakest aspect, the evaluators split between color accuracy and white balance and highlight rendering and halation---each named by two of the four---with contrast and black levels also cited; one judged the output convincing enough to change nothing.
The color-accuracy signal converges with the lower Part~1 stock scores and with the scene-adaptivity limitation of Section~\ref{sec:boundary}, and together with the highlight-rendering concern marks color fidelity and highlight roll-off as the first targets for future work.

\paragraph{Threats.}
The panel of four admits no statistical inference, and one evaluator serves on the thesis committee, a potential source of bias.
Part~1 collects absolute match ratings without a competing method, so it certifies perceived fidelity but not superiority over a baseline; the discrimination in Part~2 carries the comparative weight.
The five dropped stimuli were missing uniformly across evaluators and do not bias between conditions.
A larger panel with per-condition confidence intervals is the natural next step.

\section{Discussion}
\label{sec:discussion}

\subsection{Weight Collapse as a General Failure Mode}

LUT weight collapse is not a peculiarity of our architecture.
It is a predictable consequence of softmax-gated mixture models trained end-to-end with differentiable rendering backends.
When each expert (\ie, each basis LUT) is optimized alongside the gating network, the system reduces training loss most efficiently by specializing a single expert and routing all inputs to it---the expert collapse that load-balancing losses were designed to counter in large mixture-of-experts models~\cite{shazeer2017moe}, and a close cousin of mode collapse in GANs.

Our mitigation---temperature scaling plus entropy regularization---is lightweight.
The learnable $\tau$ is a single scalar; the entropy term adds one line to the loss computation.
The ablation demonstrates that this minimal intervention restores full weight diversity and recovers about 1~dB of PSNR.
We expect this finding to generalize to other mixture-of-experts architectures in image processing where expert modules are initialized similarly and trained from scratch.

The residual scale collapse observed in StyleLUTNet training (Section~\ref{sec:stylelut_training}) reinforces the broader lesson: auxiliary regularization losses in LUT-based architectures must be carefully balanced against the primary objective.
In both cases---softmax weight collapse and residual scale collapse---the optimizer exploited a degree of freedom (weight distribution or scale parameter) to minimize a secondary loss at the expense of the reconstruction task.

\subsection{Why a Single Conditional LUT Suffices}
\label{sec:single_lut}

The SwinLUT experiments demonstrated that five basis LUTs with entropy regularization achieve 22.16~dB on FiveK Expert~C.
StyleLUTNet, using a single conditional LUT predicted directly from the reference, achieves 22.05~dB on a different (self-supervised) evaluation protocol.
These numbers are not directly comparable---different tasks, different ground truths---but the proximity suggests that the primary benefit of multiple basis LUTs lies in increasing effective LUT resolution rather than enabling qualitatively different color mappings.

A conditional single-LUT architecture has a decisive practical advantage: it generalizes to arbitrary references without retraining.
The multi-LUT approach requires a fixed set of target styles defined at training time.
For film emulation, where users expect to point the system at any film scan and obtain a matching look, open-set capability is non-negotiable.

\subsection{Why the Final Encoder Is a CNN}
\label{sec:cnn_choice}

The Swin-T encoder was adopted to probe the multi-LUT paradigm at full strength (Section~\ref{sec:swinlut}); the final model retires it, and each reason is experimental.
First, the failure that motivated a strong backbone turned out to be backbone-independent: removing entropy regularization collapses the gate outright regardless of encoder (Table~\ref{tab:ablation}), so encoder strength buys no protection.
Second, treating the collapse exposes a capacity mismatch: 28.3M encoder parameters fund five scalar weights, and on 3{,}150 training images that capacity surfaces as a 3.76~dB train--test gap (V2), which heavier regularization only pares to 2.79~dB while test PSNR plateaus (22.19 to 22.16~dB, Table~\ref{tab:comparison}).
Third, the encoder's task shrinks in the conditional design: a global color operator requires a summary of global color statistics, not the long-range spatial modeling that attention provides (Section~\ref{sec:stylelut}).
StyleLUTNet allocates capacity accordingly---an 11.2M ResNet-18 summarizes the reference while the 71.4M decoder carries the LUT output---and reaches equivalent reconstruction quality on its own protocol (22.05 vs.\ 22.16~dB, Section~\ref{sec:single_lut}) with the open-set generalization the bank cannot offer.

\subsection{Threats to Validity}
\label{sec:threats}

\paragraph{Construct validity.}
The evaluation pairs are synthesized by random color transforms, not photographed through emulsion.
They test whether the model inverts color distortions---the StyleLUTNet objective---but do not certify that a recovered LUT matches a specific stock's spectral response.
FilmAnalyzerNet is likewise trained on synthetic film looks produced by our own renderer, so it recovers the renderer's parameters, not physically measured ones.
Processing is also display-referred throughout---operating on values encoded for display, as sRGB is, rather than proportional to scene light: the pipeline models neither the standardized encode/decode transfer functions between light and code values (OETF/EOTF) nor the log-density domain in which film response is physically specified.
The reference scans compound this: negative and reversal stocks are digitized under different color-management conventions~\cite{giorgianni2008dcm} (scene-colorimetry reconstruction for negatives, \ie, inverting to an estimate of the original scene's colors; a metameric match for reversal stock, reproducing its as-viewed appearance), so the extracted look entangles emulsion design with the scanner's rendering choices.

\paragraph{Internal validity.}
SwinLUT is evaluated on FiveK Expert~C and StyleLUTNet on the self-supervised protocol, so their PSNRs (22.16 vs.\ 22.05~dB) are not a controlled comparison.
The proximity argument of Section~\ref{sec:single_lut} is suggestive, not conclusive; a same-protocol head-to-head would settle whether one conditional LUT matches a five-LUT bank.

\paragraph{External validity.}
Open-set generalization is shown qualitatively across the stocks in Figures~\ref{fig:qualitative} and~\ref{fig:film_matrix}.
We report no quantitative open-set benchmark, because none with reference--target film pairs exists; generalization beyond the tested stocks and scenes remains unmeasured, and the boundary cases (Fig.~\ref{fig:boundary_cases}) mark where it degrades.
The pipeline further assumes a clean digital input.
A 3D LUT remaps colors pixel-wise, so uniquely digital spatial artifacts---rolling-shutter PWM banding (temporal aliasing between LED flicker and row-by-row CMOS readout), moir\'e, compression blocking---pass through the transform unchanged and can betray the digital origin of an otherwise convincing result.

\paragraph{Statistical reporting.}
Entropies reported as $0.000$ are rounded summaries of near-degenerate distributions; the measured V1 collapse has mean entropy 0.16, not exactly zero (Section~\ref{sec:swinlut_dev}).
Reported metrics are single-run point estimates without confidence intervals.

\subsection{Limitations of Pixel-Level Metrics}
\label{sec:metric_limits}

The pipeline ablation (Table~\ref{tab:pipeline}) shows PSNR decreasing as grain and halation are added.
The decrease is intended, not a regression: grain and halation are deliberate distortions that raise perceptual realism while moving pixels away from the clean reference.
PSNR penalizes any such deviation, which makes it an unreliable proxy for aesthetic quality here.
SSIM~\cite{wang2004ssim}, which compares local luminance, contrast, and structure rather than raw pixel error, is more stable across the optical stages---it holds near 0.923 while PSNR falls---confirming that grain and halation preserve structure even as they add high-frequency content.
LPIPS tracks the color stage well (the 40\% reduction from the naive baseline reflects genuine perceptual improvement) but inverts on grain, rising because its feature extractor was not trained to separate film grain from unwanted noise.
No single reference-based metric scores the full pipeline correctly; the expert evaluation (Section~\ref{sec:user_study}) supplies the missing perceptual judgment.

\subsection{Boundary Cases}
\label{sec:boundary}

\begin{figure*}[t]
  \centering
  \includegraphics[width=\textwidth]{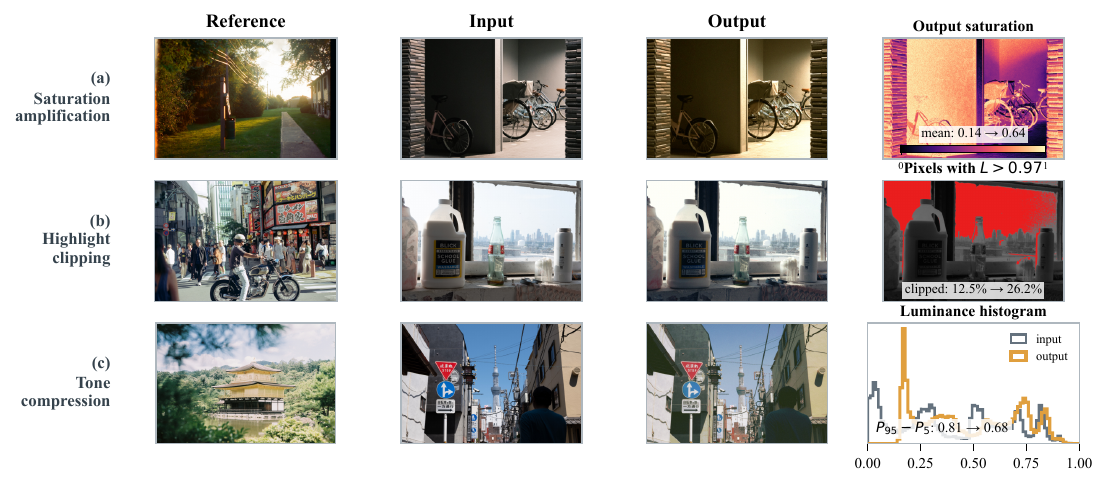}
  \caption{Boundary cases selected from the 35 open-set combinations. Columns show the reference, input, output, and selection diagnostic. \textbf{(a)} Saturation amplification. \textbf{(b)} Highlight clipping, with luminance above 0.97 in red. \textbf{(c)} Tone compression, measured by the 5th--95th percentile luminance range. Statistics use complete, uncropped images.}
  \label{fig:boundary_cases}
\end{figure*}

Figure~\ref{fig:boundary_cases} reports three boundary cases selected automatically from the 35 input--reference combinations in Figure~\ref{fig:film_matrix}.
Rather than choosing examples by visual inspection, we rank all outputs by three full-frame indicators: increase in mean saturation, increase in the fraction of pixels above luminance 0.97, and reduction of the 5th--95th percentile luminance range.
These cases are not uniformly objectionable---strong color and compressed tone can be valid film characteristics---but they expose conditions under which reference-derived statistics can dominate the input scene.

The Fujifilm reference in Figure~\ref{fig:boundary_cases}(a) raises mean saturation from 0.14 to 0.64, producing a strong warm-green cast in the low-saturation corridor scene.
In (b), the bright window scene already contains clipped highlights, but processing increases the clipped-pixel fraction from 12.5\% to 26.2\%.
In (c), the low-contrast Edo reference reduces the input's 5th--95th percentile luminance range from 0.80 to 0.68.
The three cases share a root cause: the transfer is not scene-aware.
Reference statistics encode the illumination of the reference capture---its contrast, time of day, and light quality---alongside the stock's rendering character, and the model bakes both into inputs whose lighting may differ radically.
Expert review of the deployed demo reproduced the failure: a look derived from a low-contrast interior frame produced implausible results on a high-contrast outdoor scene.
These observations identify four concrete extensions: gamut-aware constraints on the predicted LUT, highlight-preserving tone matching, an adaptive transfer-strength control based on the distance between input and reference statistics, and a scene-adaptive conditioning module that estimates the input's illumination before setting the transfer.

\subsection{Deployment}

\begin{figure}[t]
  \centering
  \includegraphics[width=0.88\columnwidth]{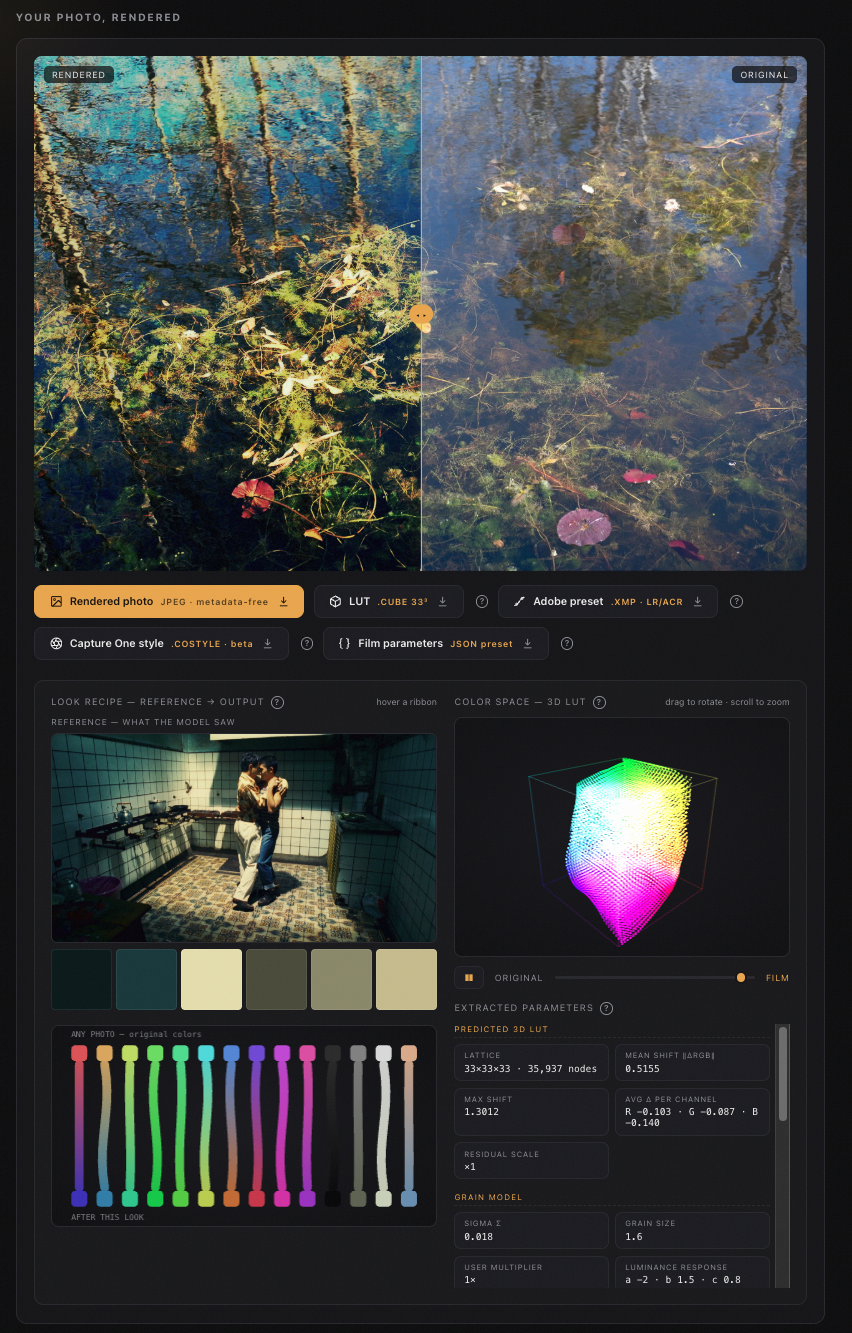}\\[2pt]
  \includegraphics[width=0.88\columnwidth]{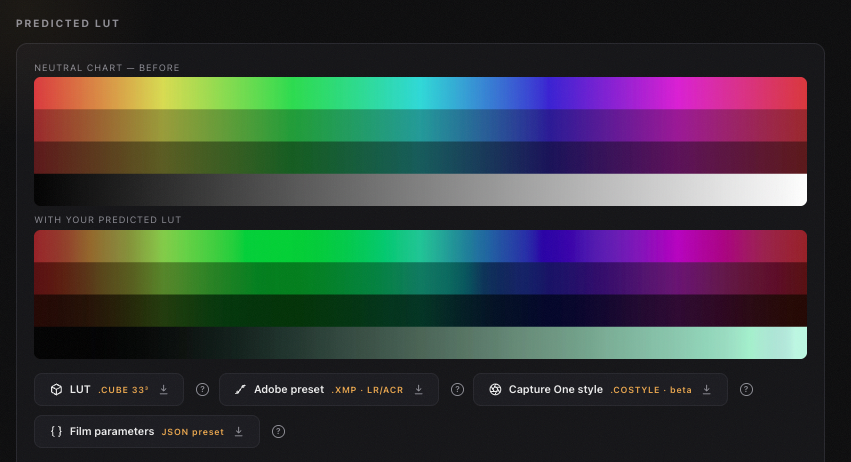}
  \caption{The public web application. \textbf{(a, top)} With a target photograph, the rendered/original comparison sits above the export row (\texttt{.cube} LUT, Adobe \texttt{.xmp}, Capture One \texttt{.costyle}, JSON parameters); below, the look recipe shows the reference as the model saw it, its extracted palette, per-hue color ribbons, an interactive rendering of the predicted $33^3$ LUT, and the regressed grain parameters. \textbf{(b, bottom)} Reference-only mode previews the predicted LUT on a neutral color chart, with the same export row. Exposing these intermediates in the browser makes each pipeline stage inspectable.}
  \label{fig:webapp}
\end{figure}

The system is publicly deployed as a web application on Hugging Face Spaces,\footnote{\url{https://huggingface.co/spaces/EtonMu/deep-analog} (accessed July 2026).} with a FastAPI backend serving the PyTorch inference pipeline and a browser-based frontend; the source is available on GitHub.\footnote{\url{https://github.com/EtonMu/deep-analog} (accessed July 2026).}
Users upload a reference film scan---the target photograph is optional, so a reference alone yields a downloadable preset---adjust three rendering parameters (tone curve strength, film tone strength, grain multiplier), and receive the rendered image together with the exportable assets: the predicted \texttt{.cube} LUT, an Adobe \texttt{.xmp} preset, a Capture One \texttt{.costyle}, and the extracted parameters as JSON.
Figure~\ref{fig:webapp} shows both modes.
Beyond the rendered/original comparison, the interface exposes the quantities that Sections~\ref{sec:stylelut} and~\ref{sec:filmanalyzer} predict: an interactive rendering of the predicted $33^3$ LUT, the reference palette, per-hue color ribbons, and the regressed grain and halation parameters.
In reference-only mode, the predicted LUT is previewed on a neutral color chart before export (Fig.~\ref{fig:webapp}(b)).
Images are processed in memory and never stored.

The \texttt{.cube} export is the most practically impactful feature.
A photographer can process a single image through the web application, download the resulting LUT, and apply it to an entire shoot of hundreds of images in Lightroom or Capture One---no neural network required beyond the initial inference.
This amortizes the computational cost across an arbitrarily large batch and integrates with existing professional workflows without requiring software changes.

\section{Conclusion}
\label{sec:conclusion}

We began from a structural observation: image-adaptive multi-LUT enhancement is a softmax-gated mixture of experts, and trained end-to-end it collapses onto a single expert.
Entropy regularization---the enhancement-setting analogue of mixture-of-experts load balancing---is the decisive countermeasure, recovering about 1~dB of PSNR with a one-line loss modification.
A second degeneracy, residual-scale collapse in conditional LUT training, shares the root cause and fixes the operating principle: auxiliary regularization must stay subordinate to reconstruction.

The diagnosis points past the fixed basis.
Because a collapse-free bank is still closed-set, we predict a single 3D LUT from a reference image (StyleLUTNet), trained by self-supervision, which transfers arbitrary film looks open-set without paired data.
Around this color backbone, Deep Analog adds a physics-informed optical branch---grain and halation rendered from parameters FilmAnalyzerNet regresses from the reference---separating photometric processing from optical simulation so the result exports as a portable preset.

The full pipeline achieves 21.72~dB PSNR; its color-grading stage runs in 5.2~ms at 1080p---fast enough for interactive editing---and exports to the industry-standard \texttt{.cube} format for integration with existing photographic workflows.
The remaining latency concentrates in an unoptimized CDF tone-matching stage that a compiled kernel would remove (Section~\ref{sec:efficiency}).
The combination of open-set generalization, physically grounded rendering, and portable preset output distinguishes Deep Analog from both closed-set LUT enhancement methods and general-purpose style transfer approaches.

\paragraph{Limitations and future work.}
The current system processes images independently; video would require temporal consistency constraints on grain and halation.
The extracted look also carries the illumination of the reference capture; Section~\ref{sec:boundary} quantifies this scene-unawareness and sketches the scene-adaptive conditioning that would counter it.
The optical branch is likewise incomplete.
It renders grain and halation but not the film's modulation transfer function (MTF)---the spatial-frequency response governing fine-detail contrast---which shapes perceived film structure at least as strongly as grain; and it synthesizes grain over whatever sensor noise the input already carries rather than composing the two in variance space.
A frequency-response stage and noise-aware grain synthesis would close both gaps.
Two revisions cut deeper.
Processing is display-referred throughout (Section~\ref{sec:threats}); reformulating the pipeline in a scene-referred linear or log-density domain, with explicit treatment of encoding transfer functions, would tie recovered parameters to physically meaningful quantities.
And the route past the synthetic protocol's spectral limits---training on large-scale collections of real film scans with known stock metadata---requires modeling the scanning stage explicitly~\cite{giorgianni2008dcm}, since negative and reversal stocks are digitized under different color-management conventions and stock-specific grain and halation correlations must be learned from the scans themselves.
The long-term direction is hybrid: coupling the conditional LUT with photochemical models of specific emulsions, moving from open-set approximation of scans toward high-confidence simulation of named stocks.
The reference encoders---an ImageNet-pretrained ResNet-18 for StyleLUTNet, Swin-T for the SwinLUT probe---were chosen as light, sufficient feature extractors; a modern self-supervised ViT foundation model (e.g., DINOv2 or CLIP) would supply stronger transferable reference features and is a natural upgrade, orthogonal to the LUT parameterization and self-supervised objective that constitute the contribution.
Finally, the decoder MLP accounts for the bulk of StyleLUTNet's 82.6M parameters; distilling it into a lighter predictor would enable on-device inference for mobile photography.


\end{document}